\documentclass[12pt]{article}
\usepackage[toc,page]{appendix} 
\usepackage{amsmath}
\usepackage{graphicx}
\usepackage{booktabs}
\usepackage{caption}
\usepackage{lineno,hyperref}
\usepackage{multirow}
\usepackage[outdir=./]{epstopdf}
\usepackage[export]{adjustbox}
\usepackage{subfig}
\usepackage[margin=1in]{geometry}
\usepackage{siunitx}
\usepackage{array}
\usepackage{subfig}
\usepackage{color, colortbl}
\usepackage[section]{placeins}
\usepackage{newunicodechar}
\usepackage[utf8]{inputenc}
\usepackage{textcomp}
\usepackage{xcolor}
\usepackage{algorithmic}
\usepackage{pdfpages}

\usepackage{hyperref}
\usepackage{xcolor}

\hypersetup{
    colorlinks=true,
    linkcolor={blue!50!black},
    citecolor={blue!80!black},
    urlcolor={blue!50!black}
}
\makeatletter
\def\p@figure{\color{red!60!black}}
\makeatother

\providecommand{\keywords}[1]
{
  	
  \textbf{\textit{Keywords---}} #1
}

\newboolean{showcomments}
\setboolean{showcomments}{true}

\ifthenelse{\boolean{showcomments}}
  {\newcommand{\nb}[3]{
  {\color{#2}\small\fbox{\bfseries\sffamily\scriptsize#1}}
  {\color{#2}\sffamily\small$\triangleright~$\textit{\small #3}$~\triangleleft$}
  }
  }
  {\newcommand{\nb}[3]{}
  }

\newcolumntype{L}[1]{>{\raggedright\arraybackslash}p{#1}}

\begin{document}
\newcommand{\HRule}{\rule{\linewidth}{0.5mm}} % Defines a new command for the horizontal lines, change thickness here
%---------------------------------------------------------
%	HEADING SECTIONS
%---------------------------------------------------------

% \include{cover_page}
% \include{title_page}

%---------------------------------------------------------
%	TITLE SECTION
%---------------------------------------------------------

% \title{\textbf{Detection of Adrenal anomalous findings using  spinal CT images }\\[0.3in]}
\title{\textbf{Detection of Adrenal anomalous findings in spinal CT images using multi model graph aggregation }\\[0.3in]}

%---------------------------------------------------------
%	AUTHOR SECTION
%---------------------------------------------------------

\author{
  Carmel, Shabalin\\
  Software and Information\\Systems Engineering\\
  Ben-Gurion University\\
  \texttt{carmelshablin@gmail.com}
  \and
  Israel, Shenkman\\
  Department of Diagnostic\\Imaging Soroka Medical Center\\
  \texttt{shenkman.israel@gmail.com}
  \and
  Ilan, Shelef\\
  Department of Diagnostic\\Imaging Soroka Medical Center\\
  \texttt{ilans@clalit.org.il}
  \and
  Gal, Ben-Arie\\
  Department of Diagnostic\\Imaging Soroka Medical Center\\
  \texttt{galben@bgu.ac.il}
  \and 
  Alex Geftler. M.D.\\
  Ortopedic Surgery Department\\Soroka Medical Center\\
  \texttt{ageftler@gmail.com}
  \and
  Yuval, Shahar\\
  Professor at\\
  Ben-Gurion University\\
  \texttt{yshahar01@gmail.com}
}
      
% \author{
%         Carmel Shabalin \\
%         E-mail: carmelshablin@gmail.com\\
%         % Department of Software and Information Systems Engineering\\
%         Ben-Gurion University of the Negev
%         }

      %   \alignauthor Roni Stern\\
      % \affaddr{Professor at Ben-Gurion University}\\
      % \affaddr{Ben-Gurion University}\\
      % \affaddr{Be’er Sheva, Israel}\\
      % \email{sternron@post.bgu.ac.il}}

\maketitle

% \vfill
% \date{\today}

%---------------------------------------------------------
%	LOGO SECTION
%---------------------------------------------------------
%\vfill
% \begin{center}
% \newcommand*{\plogo}{\includegraphics[width=0.25\textwidth]{Pictures/BGU_lable.png}}
% \plogo\\[1cm] % Include a department/university logo - this will require the graphicx package
% \end{center} 
%---------------------------------------------------------

% \begin{center}
% \HRule \\[0.6cm]
% { \huge \bfseries Paper}\\[0.4cm] % Title of your document
% \HRule \\[1.0cm]
% \end{center}

% \textbf{Supervisor:}\\[0.2in]
% Prof. Yuval Shahar\\
% Department of Software and Information Systems Engineering\\
% Ben-Gurion University of the Negev\\
%         E-mail: yshahar@bgu.ac.il\\[1.6in]

% \begin{center}
% \textbf{Confirmation Report in Fulfilment of Confirmation of PhD Candidature}\\
% \end{center}
\clearpage
\hypersetup{linkcolor=black}
% \tableofcontents
% \clearpage

\section*{ABSTRACT}
\label{scn:Abstract}
Low back pain is the symptom that is the second most frequently reported to primary care physicians, effecting 50-80 percent of the population in a lifetime ~\cite{pokrasdiagnosis}, resulting in multiple referrals of patients suffering from back problems, to CT and MRI scans, which are then examined by radiologists. The radiologists examining these spinal scans naturally focus on spinal pathologies and might miss other types of abnormalities, and in particular, abdominal ones, such as malignancies. Furthermore, the time limitation Radiologists have for each analysis severely limits their capability for considering additional diagnoses. However, the abdominal area appears in most of the scans, although it is rarely the focus of radiologists examining the spinal scans. Nevertheless, the patients whose spine was scanned might as well have malignant and other abdominal pathologies. Thus, clinicians have suggested the need for computerized assistance and decision support in screening spinal scans for additional abnormalities. 

Although the motivation, the methodology, and the results of this thesis are quite general, We have focused in the current study on the detection of one particular type of abdominal lesion that might be malignant - potentially suspicious lesions in the adrenal glands. 
There are two adrenal glands in our body, one above each kidney. There are also several types of malignancies that might develop in the adrenal glands.
In the current study, We have addressed the important case of detecting suspicious lesions in the adrenal glands as an example for the overall methodology we have developed.

% new pragraph Computational METHODOLOGY
A patient's CT scan is integrated from multiple slices (images) with an axial orientation.  Our method determines whether a patient has an abnormal adrenal gland, and localises the abnormality if it exists. 
Our method is composed of three deep learning models; each model has a different task for achieving the final goal.
We call our compound method the M\textit{ulti Model Graph Aggregation }(MMGA) method. The architecture's pipeline composition includes:
\begin{enumerate}
\item 
A CNN model based on an architecture proposed by Ben-Haim ~\cite{ben2022deep}, predicting for each slice whether it is located within the region of the adrenal glands along the Axial axis. This method filters slices of interest.
\item 
An object detection model based on the YOLO V3 architecture ~\cite{redmon2018yolov3}, which detects potentially anomalous lesions in the region of the adrenal glands, within every slice in the patient's scan.
\item 
A Graph CNN model based on the GDCNN architecture ~\cite{wang2019dynamic} aggregating the entire patient's anomalous Adrenal classifications from the previous step, into a single probability value predicting whether the patient has an anomalous adrenal gland.
\end{enumerate}

% \Carmel{TODO: A BRIEF DESCRIPTION OF THE EVALUATION Methodology}

The evaluation of the performance of our proposed method is composed of assessing two aspects, \textit{classification} and \textit{localization}. 
The \textit{classification} task's objective is to categorize a CT scan into one of the following groups - '\textit{Abnormal Adrenal found}' or '\textit{no abnormal Adrenal found}'; the method outputs a probability for a scan to have an abnormal adrenal gland. 
The \textit{localization} task can be described at two levels: First, the method provides the scan's slice index in which the abnormal adrenal gland is detected; Second, our method outputs an object localization bounding box as a description for an abnormal Adrenal found in that slice/image. The task is to localize the abnormal adrenal glands (left and right) as closely as possible to the ground truth bounding box.

The results of the evaluation show that:

\begin{enumerate}
\item Classification
\begin{enumerate}
    \item 
    Although the prior probability of a suspicious lesion in the adrenal in our data set is only 9.76\%, we have reached a positive predictive value [PPV] of up to 67\% -- an almost 7-fold higher risk for a truly suspicious lesion, compared to the background prevalence.
    
    \item 
    On the other hand, although the prior probability of a patient NOT having an abnormality is 90.24\%, our negative predictive value [NPV] is up to 94.4\%, thus reducing the risk of an abnormality, when our system says that the adrenal seems normal, from 9.76\%  down to 5.6\%, i.e. , a relative reduction of the risk by around 42\%.
\end{enumerate}
\item Localization
\begin{enumerate}
    \item 
    The average error for slice index prediction is: 8.2 for the left Adrenal and 2.25 for the right Adrenal. (each  error unit corresponds to one slice being further away from an actual slice index in which the abnormal adrenal is found; a perfect score is 0)
    \item 
    The average Intersection Over Union [IOU] score is: 0.41 for the left Adrenal and 0.52 for the right Adrenal. (a perfect score is 1.0)
\end{enumerate}

\end{enumerate}

The novelty in this study is twofold.

First, the use, for an important screening task, of CT scans that are originally focused and \textbf{tuned} for imaging the spine, which were acquired from patients with potential spinal disorders, for detection of a totally different set of  abnormalities such as abdominal Adrenal glands pathologies. 

Second, we have built a complex pipeline architecture composed from three deep learning models that can be utilized for other organs [such as the pancreas or the kidney], or for similar applications, but using other types of imaging, such as MRI.

\hspace{10pt}
\keywords{Convolutional neural networks, Graph Neural Network, aggregation with Graphs, image aggregation, diagnosis, spinal scans, Scout images of spine, abdominal pathology, Anomaly detection, Screening, adrenal glands}

\clearpage

\section{INTRODUCTION}
\label{scn:Intro}
% pancreas- lavlav,
% Kidneys - klayot,
% liver - kaved,
% Gallbladder - kis mara

\subsection{Motivation}

Cancer remains a significant public health concern worldwide, being the second leading cause of death in the US in 1997 ~\cite{greenlee2000cancer}. Early detection of tumors in CT and MR scans can greatly benefit patients. For instance, hepatocellular cancer (HCC) symptoms often appear late, but early diagnosis allows for potentially curative treatments ~\cite{singal2014early}. Similarly, kidney cancer prognosis is heavily dependent on early detection, with localized stage presenting a 93\% 5-year survival rate compared to just 12\% for distant stage ~\cite{sung2021trends}.
In recent years, there has been a rapid increase in CT scans performed in the US and UK ~\cite{hall2008cancer}. Magnetic resonance (MR) imaging plays a crucial role in patient evaluation and treatment selection ~\cite{sala2013added}. However, the expanding amount of radiological information poses challenges for interpretation.
Computer-aided diagnosis (CAD) has been proposed as a solution to assist radiologists. CAD can help reduce inter-observer variability and shorten routine steps, resulting in fewer mistakes caused by fatigue ~\cite{summers2003road, chmelik2018deep}. CAD utilizes various radiological information, including Computed Tomography (CT) and Magnetic Resonance Imaging (MRI).
While MRI provides better contrast in soft-tissue imaging, CT is faster, cheaper, and more widely accessible. In a study of twenty-six patients with primary tumors of bone or somatic soft tissues, MRI was the preferred imaging method for assessing the extent of bone and soft-tissue tumors ~\cite{aisen1986mri}.
There's significant room for improvement in CAD systems, as radiologists still face a heavy workload and diagnostic errors. An average of 3-5 percent of radiological diagnoses contain errors or lacks ~\cite{brady2017error}. 
Our proposed method uses CT scans for identifying anomalous Adrenals but can be extended to other types of scans like MRI and to detecting anomalies in other abdominal organs. This system could significantly reduce the workload on radiologists and improve detection rates of anomalous organs.
A study of 400 consecutive adult outpatients (mean age: 49 years) undergoing lumbar spine CT for low back pain found that extraspinal findings were present in 162 (40.5\%) examinations, with 4.3\% of the scans showing clinically important findings of early-stage renal cell carcinoma, transitional cell carcinoma, and other serious conditions ~\cite{lee2012extraspinal}.

\subsection{Some reaserch background}
This reaserch is in the field of Radiology pathology detection concentrating in the abdominal area in extra spinal scout images, specifically in Adrenal glands. We use CT scans of patients with back illnesses for anomalous Adrenal glands detection. The scanning performed at Soroka Medical Center, in collaboration with the head of the MEDICAL IMAGING institute. 
The expert Radiologists label the scans for normal Adrenals and for abnormal/anomalous Adrenals. the process includes localizing each of them within the scan by a bounding box if exist. These bounding boxes are used as the ground truth for training our models. Some percentage of the labeled data will be put aside for final testing our models and calculating accuracy and other metrics such as receiver operating characteristic (ROC) curve, precision/PPV and NPV. 

CT scans provided form Soroka Medical Center are used as inputs for our method. Each patient's scan consists of multiple ordered images or slices, the order is corresponding to the physical height in Axial axis.
We use Deep Learning (DL) methods particularly Convolutional Neural Networks (CNN) to build a pipeline method for detecting and localizing anomalies in the Adrenal glands, CNN has brought about breakthrough in pattern recognition of images including medical images \cite{kido2018detection}.

\subsection{Methodology and Novelty}
This research examine different Deep Learning methods of supervised learning with emphasis on a Medical CT Scans as an input data. The aim of these methods is to detect and classify abdominal cancerous pathologies.
The novelty of our reaserch is twofold - first, utilizing images taken from patient with spinal illnesses or back pain to detecting completely other non-related to spine illnesses. CT scan images are used, and the focus of these scans are the spine i.e. the CT operator is setting CT properties making the spine appear clearly in resulting scan, as a result the extra spinal area our region of interest in the images is less clear, making it harder to detect abnormalities in that area.

Second, MMGA method architecture can be utilized for using other image types such as MRI and also detecting verity of abdominal anomalies. The pipeline method constructed from 4 consecutive steps each step solving different problem implemented by different DL model:
\begin{enumerate}
\item 
% \norm{f}_{L_{2}(\Omega)}
In this step we filter slices of interest, those slices where the Adrenal glands are found or close to in Axial axis. We build a CNN model based on architecture proposed by Ben-Haim ~\cite{ben2022deep} predicting each slice whether it is located near region of Adrenals along the Axial axis. 
\item 
The input of this steps is all patient's slices/images after previous filtering step. Anomalous Adrenal detection step, implemented using a model based on YOLO V3 architecture ~\cite{redmon2018yolov3} which detects anomalous Adrenals within every slice in patient's scan, the output of this step is bounding boxes with confidence score for every anomalous Adrenal found - left or right.
\item 
The final step is aggregating previous' step. The input of this step is patient's bounding boxes for all scan images/slices. we build Graph CNN model based on GDCNN ~\cite{wang2019dynamic} aggregating entire patient's anomalous Adrenal defections to a single value predicting whether a patient has anomalous Adrenal or not.
\item 
After final step of predicting whether a patient has anomalous Adrenal, we localise the anomalous adrenals (left or right) composing slice and the bounding boxes within the slice. we use intermediate result (these are bounding boxes for an image and confidence scores) from second step which the YOLO architecture ~\cite{redmon2018yolov3} provides and aggregate them to final slice and bounding boxes. 
\end{enumerate}

% \subsection{Implication Of The Research}
% \begin{enumerate}
%     \item  
%     Building a system that gets as input CT scans that are directed and tuned for the spine and which were taken from patients with back issues, and using it for automatically detecting different illnesses such as Adrenal pathologies, can significantly assist radiologists in interpreting spinal scans,  reduce False Negative diagnostic results, and enhance the safety of patients.
%     % \item Building a system that can harness information from multiple types of imaging scans, in particular, CT and MRI, might enhance the accuracy of detecting anomalies in the abdominal areas, compared to the use of only one imaging mode.
%     \item 
%     Is it possible to detect and classify Adrenal pathologies from CT spinal scans, using CNN networks and deep learning methods?
%     \item 
%     Creating a generic recipe for building unique 3 fold pipeline can benefit to whole other organs anomaly detection not restricted to abdominal area and CT scans.
%     \item 
%     Creating a method supporting simultaneously multiple organs detection .i.e a single model detecting both left and right anomalous adrenals.
%     \item 
%     Can final aggregation step in pipeline which predicts whether patient's entire scan has abnormal Adrenal can be simplified form Graph learning complex method to a simple rule based method achieving same or better performance.
    
% \end{enumerate}

\section{DOMAIN BACKGROUND}
\label{scn:Background}
\subsection{Abdominal pathologies and anatomy}
\label {anatomy}
\subsubsection{Basic abdominal anatomy relevant to adrenal glands }
The abdomen contains various vital organs, including the stomach, intestines, liver, pancreas, kidneys, and importantly for our study, the adrenal glands. The adrenal glands, also known as suprarenal glands, are small, triangular-shaped endocrine glands located on top of each kidney.

% \begin{figure}[!h]
%     \centering
%     \includegraphics[scale=0.7]{Pictures/Abdomens.png}
%     \caption{Picture of the Abdomen}
%     \label{fig:abdomen}
% \end{figure}

\begin{figure}[!h]
    \centering
    \includegraphics[scale=0.4]{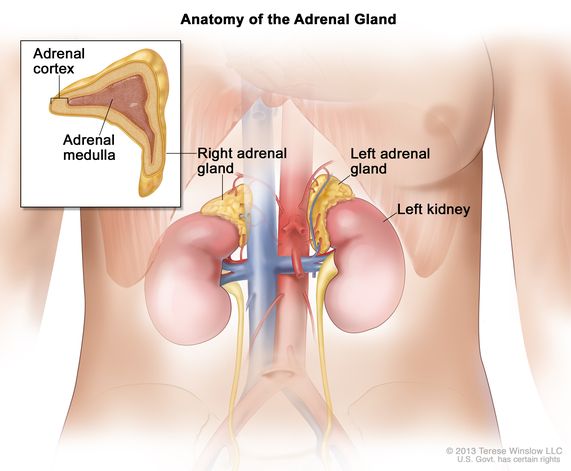}
    \caption{Adrenal glands zoom in}
    \label{fig:abdomen adrenal}
\end{figure}

Each adrenal gland consists of two distinct parts: the outer adrenal cortex and the inner adrenal medulla. The adrenal cortex produces hormones essential for metabolism, immune system function, and regulation of blood pressure. The adrenal medulla produces catecholamines, including adrenaline and noradrenaline, which play a crucial role in the body's stress response.
The location and small size of the adrenal glands make them challenging to visualize and assess using traditional imaging techniques, which is why advanced imaging and analysis methods are crucial for accurate diagnosis of adrenal pathologies.

\subsubsection{Common abdominal diseases and prevalence in population}
\label{bcg:Common abdominal diseases}
% In a prospective cohort of patients, examined abdominal symptom frequency, initial diagnostic suspicion, and actions of general practitioners (GPs) in response to abdominal symptoms that may indicate cancer. 61802 patients, Abdominal symptoms were recorded in 6264 (10.1\%) patients. A subsequent malignancy was reported in 511 patients (0.8\%): 441 (86.3\%) had a new cancer, 70 (13.7\%) a recurrent cancer~\cite{holtedahl2017abdominal}. 

In a prospective cohort study of 61,802 patients with abdominal symptoms, Abdominal symptoms were recorded in 6264 (10.1\%) patients, subsequent 511 patients (0.8\%) were diagnosed with malignancies. Of these, 441 (86.3\%) had a new cancer, and 70 (13.7\%) had a recurrent cancer ~\cite{holtedahl2017abdominal}. This data underscores the importance of accurate diagnostic tools for abdominal pathologies.

A study performed on National Italian Study Group a multicentric retrospective analysis of adrenal masses incidentally discovered (adrenal incidentalomas) concluded that occurrence of incidentally discovered adrenocortical carcinomas and pheochromocytomas is not rare \cite{angeli1997adrenal}. The frequency of adrenocortical cancer was 12\% among operated patients, while the frequency of pheochromocytoma was 10\% among operated patients. The tumor diameter was highly correlated with the risk of malignancy, as well as the CT characteristics such as density, shape and margins.

While various abdominal cancers exist, our research focuses on anomalous adrenal glands. The detection of suspicious lesions in the adrenal glands serves as an important case for our overall methodology. Adrenal abnormalities, though less common than some other abdominal pathologies, present unique challenges in detection and diagnosis due to their location and size.

\subsection{Overview on Convolutional Neural Networks}

% The history of artificial neural networks goes back to the early days of computing. In 1943, mathematicians Warren McCulloch and Walter Pitts built a circuitry system intended to approximate the functioning of the human brain that ran simple algorithms~\cite{hertz2018introduction}.
% Artificial neural networks have evolved from their biologically inspired roots to a well established means to solve a broad spectrum of engineering problems. Their embedding into modern statistics has provided the necessary theoretical foundation for challenging engineering tasks, such as advanced real time image and signal processing~\cite{goerick1997neural}.

\begin{figure}[h]
\begin{tabular}{ll}
\includegraphics[scale=0.50]{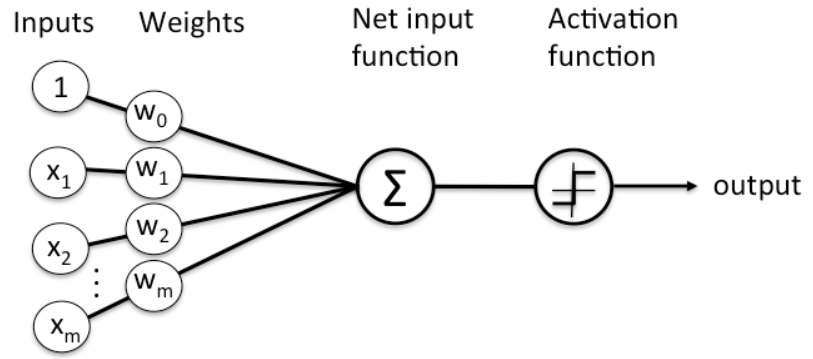}
&
\includegraphics[scale=0.95]{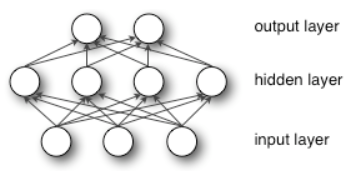}

\end{tabular}
\caption{Left : Artificial Neuron (processing unit); Right: ANN architecture}
\label{fig:Neuron}
\end{figure}

% \begin{figure}[!h]
%     \centering
%     \includegraphics[scale=0.7]{Pictures/Neuron.PNG}
%     \caption{Artificial Neuron / processing unit}
%     \label{fig:Neuron}
% \end{figure}

% \begin{figure}[!h]
%     \centering
%     \includegraphics[scale=1.2]{Pictures/NN.PNG}
%     \caption{Artificial deep neural network architecture}
%     \label{fig:ANN}
% \end{figure}
Artificial neural networks (ANN) are built like the human brain, with neuron nodes interconnected like a web. An ANN is constructed of artificial neurons also called processing units, which are interconnected by nodes and could be arranged in layers. These processing units are made up of input and output units. The input units receive various forms and structures of information based on an internal weighting system, and the neural network attempts to learn about the information presented to produce one output report usually by the last layer.

\textbf{Convolutional Neural Networks} (CNNs) have become a cornerstone in computer vision tasks, including image classification and analysis. Inspired by biological processes, CNNs are designed to automatically learn hierarchical feature representations from input data.
A typical CNN architecture consists of:

\begin{enumerate}
    \item Convolutional layers: Extract features using learnable filters.
    \item Activation layer: Introduces non-linearity, often using ReLU (Rectified Linear Unit).
    \item Pooling: Reduces spatial dimensions and computational complexity.
\end{enumerate}

\begin{figure}[!h]
    \centering
    \includegraphics[scale=1.1]{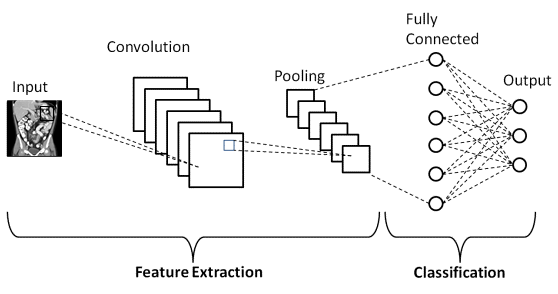}
    \caption{Convolutional Neural Network basic architecture}
    \label{fig:CNN}
\end{figure}

A classical Convolutional Neural Network demonstrated in figure \ref{fig:CNN}. Starting with input layer then a basic convolution layer and finally connecting to a fully connected layer with an output layer for classification task.

CNNs have shown remarkable success in various computer vision tasks, including medical image analysis. Their ability to learn complex patterns and features directly from raw data makes them particularly suited for tasks such as detecting anomalies in CT and MRI scans.
In the context of our research, CNNs serve as a fundamental building block for developing advanced models capable of detecting and classifying adrenal gland abnormalities in medical imaging.

\subsection{Overview on Object detection}

\begin{figure}[!h]
    \centering
    \includegraphics[scale=0.65]{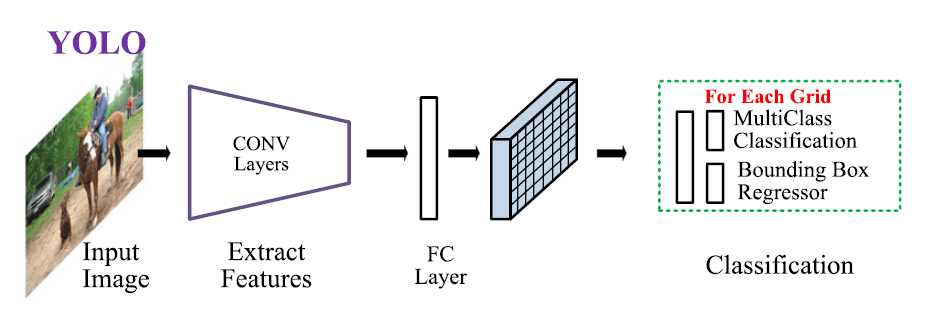}
    \caption{High level diagram of YOLO framework for generic
object detection \cite{redmon2016you} }
    \label{fig:yolo_flow}
\end{figure}

Object detection, a fundamental problem in computer vision, aims to locate and classify object instances in images. Deep learning techniques have significantly advanced this field, with YOLO (You Only Look Once) being a prominent example.
YOLO, proposed by Redmon et al.  \cite{redmon2016you}, treats object detection as a regression problem, predicting bounding boxes and class probabilities directly from full images in a single evaluation. This approach offers real-time performance, processing images at 45 FPS.

For our research on anomalous adrenal detection, we utilize YOLOv3 \cite{redmon2017yolo9000}, which improves upon the original model with better accuracy and efficiency. YOLOv3 incorporates features such as multi-scale predictions and a more powerful feature extractor network, making it well-suited for detecting small objects like adrenal glands in medical imaging.

\subsection{Overview on Graph Learning}

Graphs, representing entities as vertices and their relationships as edges, are powerful tools for modeling complex data structures. In recent years, graph learning methods have gained popularity for capturing intricate relationships in various domains, including medical imaging.
Graph learning methods map the features of a graph to feature vectors with the same dimensions in the embedding space. This approach is particularly useful for tasks such as node classification, link prediction, and graph classification.
In our research, we employ a CNN-based deep learning method called Dynamic Graph CNN (DGCNN) for learning on point clouds \cite{wang2019dynamic}. DGCNN is designed to capture local geometric structures by constructing a local neighborhood graph and applying convolution-like operations on the edges connecting neighboring pairs of points.

% \begin{figure}[!h]
%     \centering
%     \includegraphics[scale=0.16]{Pictures/EdgeConv.jpg}
%     \caption{EdgeConv layer description part of DGCNN method}
%     \label{fig:EdgeConv}
% \end{figure}

The key innovation in DGCNN is the EdgeConv operation, which dynamically computes node adjacency in the feature space, allowing the model to learn and update topological relationships throughout the network layers. This characteristic makes DGCNN particularly suitable for analyzing the complex spatial relationships in medical imaging data, such as CT scans of adrenal glands.

% Figure \ref{fig:EdgeConv} Left: Computing an edge feature,  $e_{ij}$ (top), from a point pair, $x_{i}$ and $x_{j}$ (bottom). In this example, $h_{\Theta}()$ is instantiated using a fully connected layer, and the learnable parameters are its associated weights. Right: The EdgeConv operation. The output of EdgeConv is calculated by aggregating the edge features associated with all the edges emanating from each connected vertex \cite{wang2019dynamic}.

\subsection{Computer-Aided Diagnosis}

Computer-aided diagnosis (CAD) systems, which emerged in the 1980s and 1990s in radiology ~\cite{oakden2019rebirth}, assist doctors in interpreting medical images. Modern CAD systems, powered by deep learning, have shown remarkable progress in medical imaging tasks ~\cite{xing2021artificial}. Convolutional Neural Networks (CNNs) have become fundamental in CAD applications, enabling learning of highly representative, hierarchical image features from sufficient training data ~\cite{shin2016deep}.
In the field of CAD, both 2D and 3D CNN approaches have been explored. 2D CNN architectures are computationally efficient and have shown success in various medical imaging tasks ~\cite{shin2016deep}. On the other hand, 3D CNN approaches, while more computationally intensive, can better utilize volumetric information from CT scans, but face challenges such as overfitting and increased training time ~\cite{yu2020simplified}.
Prasoon et al. ~\cite{prasoon2013deep} demonstrated the effectiveness of CNNs in medical imaging by proposing a system of three orthogonal 2D CNNs for knee cartilage segmentation, achieving high accuracy and ease of training.
These advancements in deep learning-based CAD systems offer potential for improving the detection and diagnosis of various medical conditions, including adrenal abnormalities.

\subsection{Automatic ventral anomaly detection}
There is a big challenge of automatically detecting tumors within abdominal scans, for example liver tumors show a high variability in their shape, appearance and localization. They can be either hypodense (appearing darker than the surrounding healthy liver parenchyma) or hyperdense (appearing brighter), and can additionally have a rim due to the contrast agent accumulation, calcification or necrosis. The individual appearance depends on lesion type, state, imaging (equipment, settings, contrast method and timing), and can vary substantially from patient to patient. This high variability makes liver lesion segmentation a challenging task in practice~\cite{chlebus2018automatic}.

Another challenge is the clinical efficacy of incidental findings on MRI scout images of various organs. Although most incidental findings are benign and asymptomatic, some may be related to clinically significant diseases including malignant tumors. However, the image sequence, composition, and quality of scout images differ depending on the manufacturer or institutions because they are obtained at the discretion of technologists or with the default setting of MRI without a standard protocol. Therefore, the clinical efficacy of scout images in daily practice remains debatable~\cite{you2021readability}. Currently there are no automatic extra spinal anomaly detection system, in this reaserch we utilize from these incidental findings on CT images of extra scout spinal organs and build an automatic system that can detect abnormal Adrenal glands and help Radiologists with diagnosis.

% \subsubsection{Pathologies detection in CT}

% \begin{figure}[!h]
%     \centering
%     \includegraphics[scale=0.5]{Pictures/CT_scanner.png}
%     \caption{CT scan procedure}
%     \label{fig:CT scan}
% \end{figure}
Computed Tomography (CT) is a sophisticated scanning technique used in radiology to form volumetric X-ray images of the anatomy, in the sense of tissue density maps.
CT has transformed much of medical imaging by providing three-dimensional views of the organ or body region of interest~\cite{brenner2007computed}.

% \begin{figure}[!h]
%     \centering
%     \includegraphics[scale=0.48]{Pictures/Ualike_CNN.png}
%     \caption{U-net fully convolutional network architecture}
%     \label{fig:Ucnn}
% \end{figure}

% In the field of the Liver Tumor Segmentation (LiTS) challenge Chlebus et al., \cite{chlebus2018automatic} focuses on the tumor segmentation task, which they employed a U-net alike fully convolutional network architecture shown in Fig. \ref{fig:Ucnn}. The contribution on the tumor segmentation task is by cascading of a 2D Fully convolutional neural network (FCN) working on a voxel-level with a model trained using hand-crafted features extracted on an object-level describing Liver's shape and location (e.g. liver/gallbladder boundary) which achieved state-of-the-art results in the LiTS challenge.
\begin{figure}[!h]
    \centering
    \includegraphics[scale=0.43]{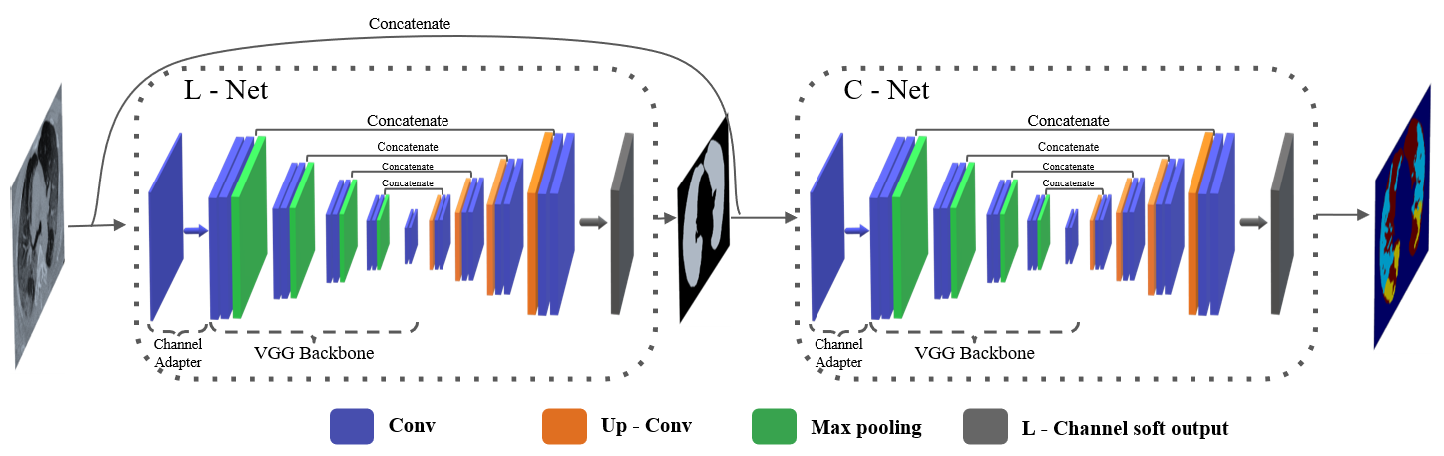}
    \caption{The compound hierarchical segmentation architecture. The left-most U-Net segments the lungs’ cavity and the right one
partitions the lung area into GGO, CON, and healthy tissue \cite{ben2022deep}.}
    \label{fig:Lung Ucnn}
\end{figure}

In the field of lung CTs segmentation Ben-Haim et al. \cite{ben2022deep} focuses on the segmentation for COVID-19 severity  assessment task,  which they employed a U-net alike fully convolutional compound networks architecture shown in Fig \ref{fig:Lung Ucnn} with the famous VGG network \cite{simonyan2014very} backbone in each U-net network. 
They achieved state-of-the-art results in lung segmentation for COVID-19 severity assessment task and provide a measure for segmentation uncertainty corresponding to the disagreements between the manual annotations of two different radiologists.

\section{METHODS}
\label{scn:Methods}

% \YUVAL {\textbf{YUVAL: EVERY "SECTION" SUCH AS "Introduction" or "Background" or "METHODS" or "Results" SHOULD BE A SEPARATE *CHAPTER* AND START ON A NEW PAGE}}

In this chapter, we describe my compound method for detecting and localizing abnormal Adrenals glands within extra spinal scout CT scan. The method steps are as follows:

\begin{enumerate}
\item 
First step goal is filtering out all "far away" slices from ROI passing to next steps. We build a CNN model predicting each slice whether it is located near region of Adrenals along the Axial axis.
\item 
Second step which detects anomalous Adrenals within every slice in patient's scan. Our model is based on YOLO V3 architecture ~\cite{redmon2018yolov3}.
\item 
A Graph CNN model based on GDCNN ~\cite{wang2019dynamic} aggregating entire patient's anomalous Adrenal detections from previous step to a single value predicting whether a patient has anomalous Adrenal.
\end{enumerate}
Next, we describe each of these steps in more detail.

\subsection{Data preprocessing}
\begin{figure}[!h]
    \centering
    \includegraphics[scale=0.25]{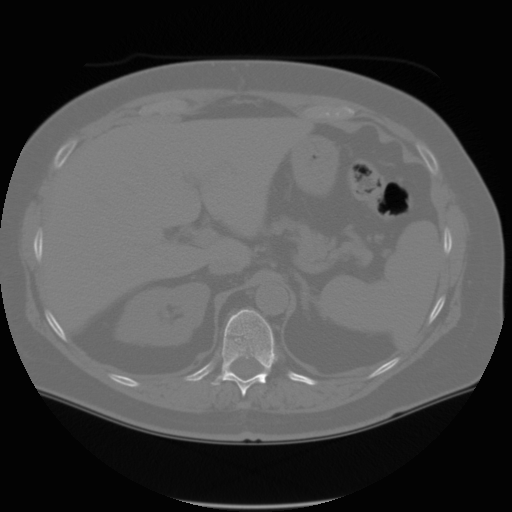}
    \caption{ Axial slice image from patient's spine CT }
    \label{fig:axial slice}
\end{figure}

The input CT scans provided by Soroka hospital orthopedic department, the scans originate from patients complaining about back issues, most scans focus on the lumbar spine and the direction of a scan is in Axial axis as shown example in Figure. \ref{fig:axial slice} a real slice from a patient scan. 

These scans provided as files in DICOM format which is the standard image file format used by radiological hardware devices \cite{graham2005dicom}. Each DICOM file represents a single slice within whole patient's scan in Axial direction, i.e. each patient has multiple slices on average about 300 slices (images) per patient. Within patient's scan, DICOM files are ordered lexicographical based on the file names, the file order is corresponding to the physical location in Axial axis from top to bottom, most scans start at Lungs area and finishing around Sacrum area. 
Each DICOM file contains a single 16 bit gray scale 2 dimensional image $X$ where $X_{i,j}$ denotes pixel in row $j$ and column $j$ the following normalization preformed on each image reducing it to 8 bit gray scale image before saving it to PNG format for later usage (train/test):

% \begin{equation}
% Normalize(X) = \frac{\max(0, X_{i,j})}{\max(X)}*255
% \end{equation}

\begin{equation}
Normalize(X) = \frac{\underset{i,j}{\operatorname{max}}(0, X_{i,j})}{\max(X)}*255
\end{equation}

After normalization and PNG conversion each scan's pixels values are numeric numbers in range of [0, 255] indicating the intensity (zero represent black, and 255 white).
The Data-set consists of 1250 patient's CT scans from the orthopedics department at Soroka University Hospital. 
Expert radiologists conducted labeling all the provided scans and classifying the type within 2 classes: normal and abnormal Adrenal classes (if exists) in the scan, and segment it's location with a bounding box on the image. This labeling are the ground truth for the learning process.

% The Data-set consists of few hundreds of patient's paired CT and MRI scans together with their metadata from the orthopedics department at Soroka University Hospital. Expert radiologists will label all the provided scans, classify the the type of carcinoma (if exists) in the scan, and detect its location with a bounding box and slice number. While classifying the scans, radiologists are blind to diagnostic and clinical results.

% The data will be heavily reprocessed in order to normalize the images, this will include shifting, rotating, flipping. Another aspect is segmenting in images the focused organ to make it easier for the model to concentrate on the relevant abdominal featured region as apposed to other informative regions. 
% Some times enough data acquiring is hard and not possible, so we might use Data augmentation techniques in order to have more data samples to train on, which can be crucial for achieving good accuracy. For example Tiexin Qin. ~\cite{qin2020automatic} developed a novel automatic learning-based data augmentation method for medical image segmentation which models the augmentation task as a trial-and-error procedure using deep reinforcement learning (DRL).

\subsection{Learning process}
% High level overview shown in Figure. \ref{fig:abdomen} describes the compound 3 step method:

\subsubsection{Step 1 - Slices of interest}  \label{step 1}
\begin{figure}[!h]
    \centering
    \includegraphics[scale=0.5]{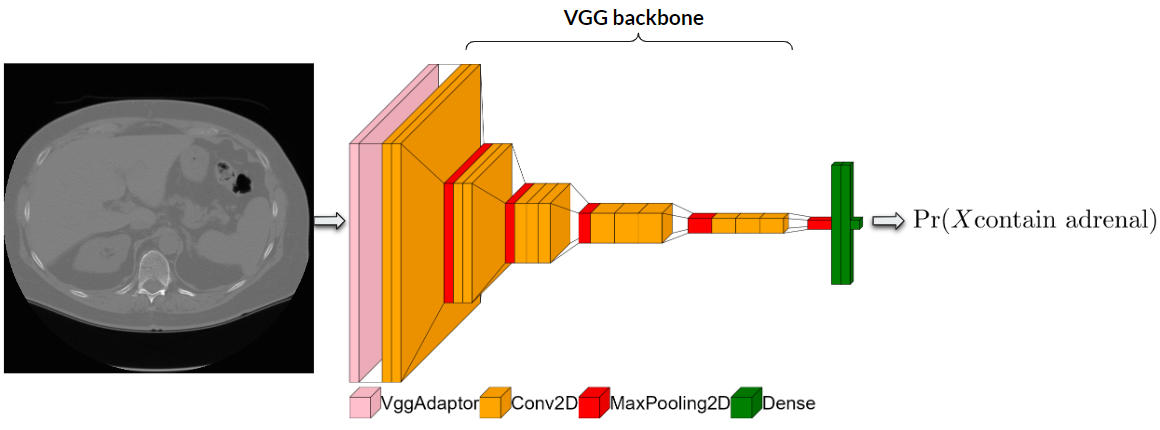}
    \caption{ Model architecture contains VGG CNN pre-trained layers finlay predicts weather an image contains Adrenal }
    \label{fig:slices model}
\end{figure}

Each patient scan contains few hundreds of ordered slices/images, starting from chest region and finishing around the pelvis. The Adrenal glands which are the objects of interest are present in small amount of consecutive slices only, the prevalence of such images are around 10\% consisting couple of tens within an entire patient's scan. The main purpose of this step is to reduce substantially the amount of images and more importantly the variability of the images to be passed for training the object detection task model. A narrow distribution of images will make it easier for the abnormal Adrenals detection model to learn and detect. The model is CNN based gets as input an image of a single slice (matrix 512 x 512 of pixels) and outputs the probability of the image to contain an Adrenal as shown in Figure \ref{fig:slices model} (third party code was used for visualization \cite{Gavrikov2020VisualKeras}).
Model's back bone build from pretrained VGG \cite{simonyan2014very} convolutional layers which was inspired by Ben-Haim et al. \cite{ben2022deep} reaserch focusing on lung CT segmentation for COVID-19 severity assessment, their first U-net shown in Fig \ref{fig:Lung Ucnn} is segmenting each pixel whether its part of the lungs or not, the similarity between the inputs and the final goal of the networks (theirs and ours) 
has led us to assume that this aspect of their architecture might be useful also for our architecture and for the current study.

\begin{figure}[!h]
    \centering
    \includegraphics[scale=0.5]{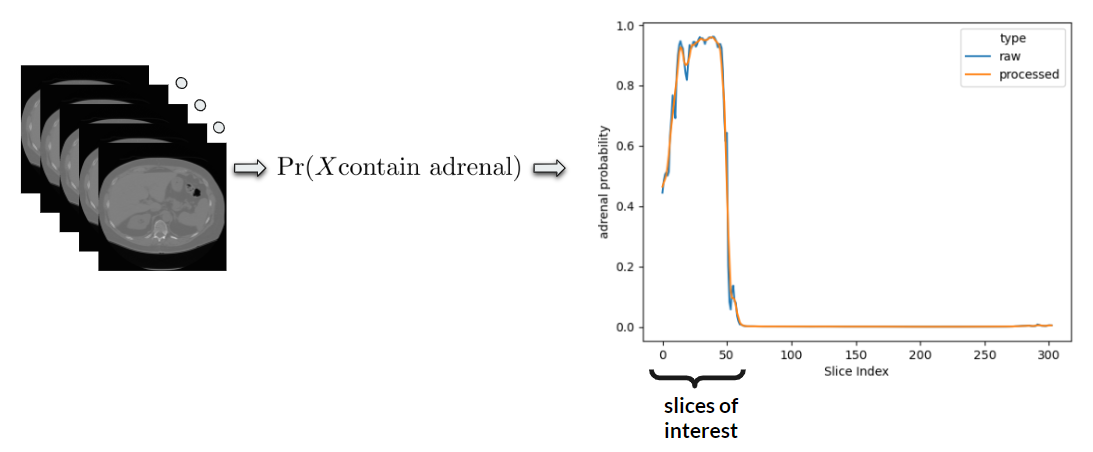}
    \caption{ Illustration of scan probabilities signal to get the slices of interest }
    \label{fig:slices proba}
\end{figure}

After all patient scan's ordered images where classified with probabilities by the model (low probability means no Adrenal in image and Vice Versa), to get the final slices of interest that contain the Adrenal glands we performed signal processing method based on moving average and threshold. Assume patient's scan has $n$ ordered images, let $x_{i}$ represent image order $i$, let $p_{i} = \Pr(x_i \text{ contain adrenal}) \forall i \in \{1..n\} $, define moving average of $p_{i}$ with window size equal to $5$:
\begin{equation}
\label{eqn:Location}
    ma_i = ma(p_{i}) = \frac{\sum_{j=i-2}^{i+3} p_j}{5} 
\end{equation}
let $proba\_th = 0.4$ is the minimum probability for an image to have in order to be considered as image containing an Adrenal, define all potential image indices with probability grater than $proba\_th$ as: $potential\_inx=\{ i \mid ma_i>proba\_th \}$, Finally all images to be considered as slice of interest to be $\{ x_{i} \}_{\min(potential\_inx)}^{\max(potential\_inx)}$;
A High level overview of scan probabilities signal shown in Figure. \ref{fig:slices proba}

\subsubsection{Step 2 - Abnormal Adrenal detections} \label{step 2}
After each scan passed through "slices of interest" method shown in section \ref{step 1} the remaining images reduce substantially the variability of image space, since the slices that were left (the precise details will be described in the Results chapter) are only slices around the small physical space near the Adrenal glands .i.e scan images above the stomach which include lungs and shoulder bones are filtered-out or images below the kidneys that include pelvic bones are filtered-out as well. Now the task of abnormal Adrenal detection is easier not only due to the high image count reduction but also because of smaller image variability. A good and very fast known CNN based architecture called YOLO \cite{redmon2018yolov3} used for abnormal Adrenal detection which models the task of classification and localization all-together. A pretrained YOLO v3 model (trained on Darknet-53 dataset \cite{redmon2018yolov3}) was used for the reaserch, performing retraining with CT scan images (with their corresponding bounding boxes as ground truth) that passing filtering method in section \ref{step 1}. The retrained YOLO model gets as input a single 2D image from single patient's scan and outputs multiple bounding boxes containing potential abnormal glands (left and right glands) with their corresponding confidence scores.

\begin{figure}[!h]
    \centering
    \includegraphics[scale=0.55]{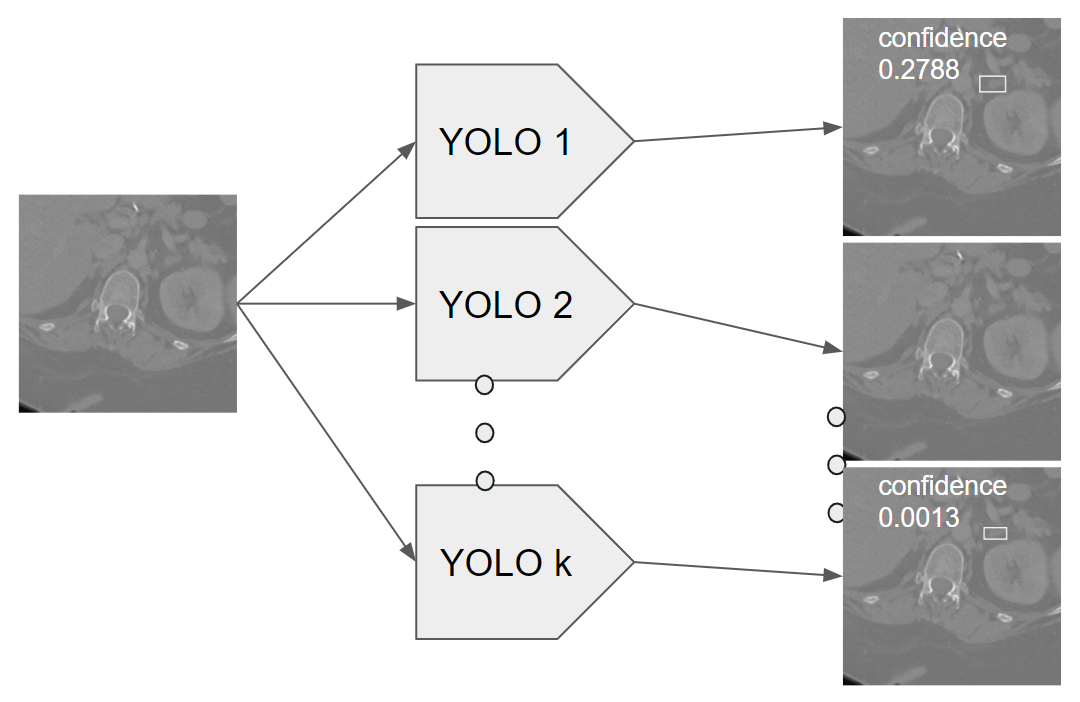}
    \caption{Trained different YOLO models detect abnormal Adrenals differently.}
    \label{fig:multi yolo}
\end{figure}

A common practice in medicine images like object detection/segmentation tasks for reducing false negative rate and increasing accuracy is applying ”wisdom of crowd” method. Ben-Haim et al. \cite{ben2022deep} utilized statistics of the ensemble independent predictions allowing enhancement the overall segmentation accuracy. In this reaserch we apply ”wisdom of crowd” method by training $k$ YOLO models, the training processes of the networks
differ due to the random weight initialization of the layers
weights as well as the random shuffling, resulting in different predictions between models for-each image such that a high true positive between models agreement on patients with abnormal Adrenal and very low agreement on patients with normal adrenals false positively predicted. This notion explain why aggregating multiple predictors beneficial to reduce false positive rate, leading to next section \ref{step 3} that explains how aggregation for all slices for all $k$ YOLO models predictions is preformed for each patient. Figure \ref{fig:multi yolo} illustrate the resulting detections for $k$ different YOLO models for a single slice taken form patient's scan, it can be seen that predictions are different between models, $YOLO_2$ model couldn't detect abnormal Adrenal while $YOLO_1$ and $YOLO_k$ detect in similar area but not identically.

\subsubsection{Step 3 - Detections Aggregation with Graph learning} \label{step 3}

At this stage we are left with abnormal Adrenal predictions from $k$ different YOLO models for a subset of consecutive images from the original patient's scan passing section \ref{step 2}.
Graph learning have been used for aggregation of aggregation in Computational Pathology,
multi-gigapixel whole-slide images (WSIs) often demand processing a large number of tiles (sub-images) and require aggregating predictions from the tiles in order to predict WSI-level labels.
The WSI representations for predictive modelling offer options from classical machine learning to graph learning where graph convolutional neural networks (GCNNs) learn and aggregate all information into a single score for clinical decision-making \cite{bilal2023aggregation}.
Similarly in this reaserch using graph learning aggregating the abnormal adrenal predicted bounding boxes into a single score of patient probability having an abnormal Adrenal(left or right Adrenal).

\begin{figure}[!h]
    \centering
    \includegraphics[scale=0.45]{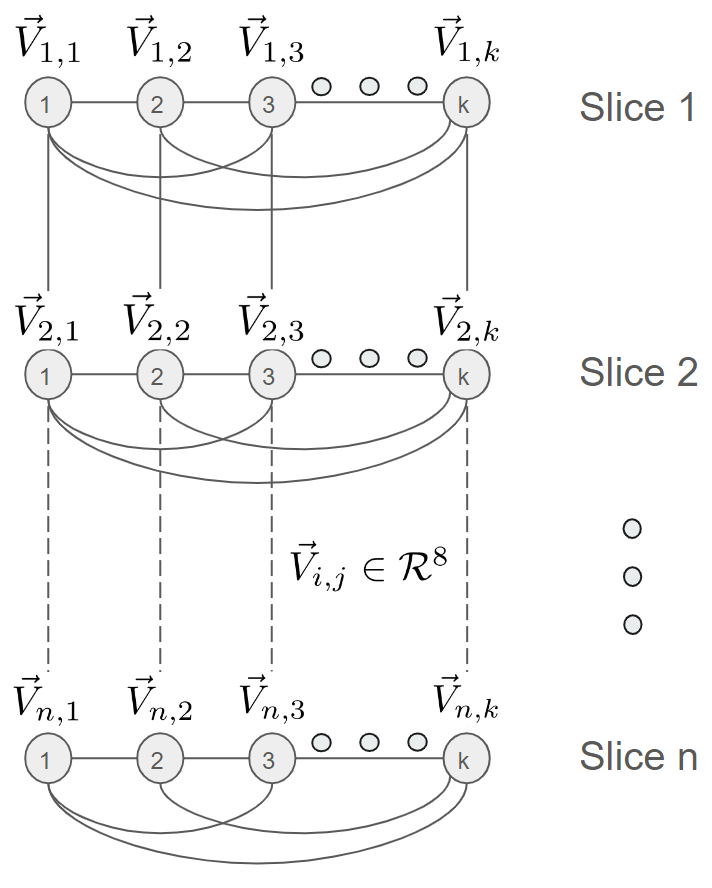}
    \caption{Patient scan transformed into a graph.}
    \label{fig:yolo graph}
\end{figure}

\textbf{Graph construction:} Each patient's scan transformed into a graph as Figure \ref{fig:yolo graph} illustrate, assume patient's scan has $n$ consecutive slices passed "Slices of interest" section \ref{step 1}, each slice/image is represented by $k$ nodes, one node for every YOLO model's predicted bounding boxes described in section \ref{step 2}, each node has 8 numerical features representing 2 bounding boxes for left and right Adrenals. The first 4 dimensions represent a left Adrenal bounding box found in the 2D image and the last 4 dimensions represent the right adrenal found. Bounding box description, let every 2D image pixel has $x$ and $y$ values, a single bounding box vector representation is as follows - $(\text{top left }x, \text{top left }y, \text{bottom right }x, \text{bottom right }y)$. If no Bounding box found for left Adrenal for example then its bounding box's feature values set to $-1$ represent a non valid pixel value (a valid pixel value is non negative).

All $k$ nodes representing a single slice are connected between themselves with edges in total $\binom{k}{2}$ edges for each slice (horizontal edges). Nodes represented by consecutive slices are connected only if they represent the same YOLO model (vertical edges).

\textbf{Graph Learning Method:} Dynamic Graph CNN (DGCNN) for Learning on Point Clouds is a recent reaserch propose a new neural network module dubbed EdgeConv suitable for CNN-based high-level tasks on point clouds, including classification and segmentation. This method adapting insight from CNN to the point cloud world. Point clouds inherently lack topological information, so designing a model to recover topology can enrich the representation power of point clouds. EdgeConv acts on graphs dynamically computed in each layer of the network, it has several appealing properties: It incorporates local neighborhood information; it can be stacked applied to learn global shape properties; and in multi-layer systems affinity in feature space captures semantic characteristics over potentially long distances in the original embedding \cite{wang2019dynamic}.

\begin{figure}[!h]
    \centering
    \includegraphics[scale=0.32]{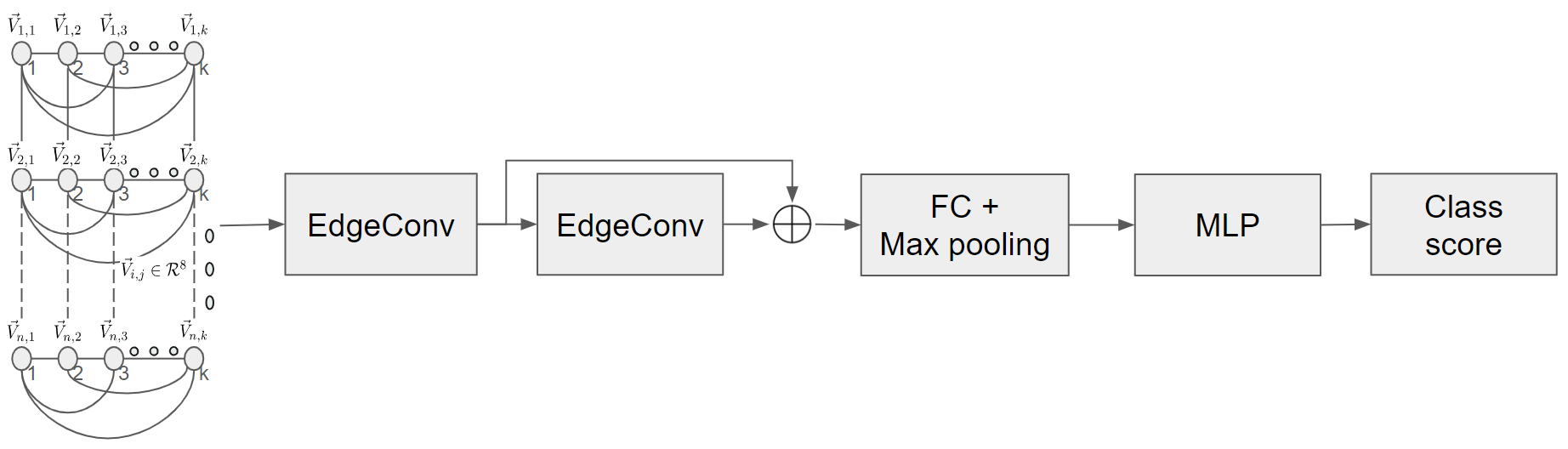}
    \caption{DGCNN architecture used for graph classification}
    \label{fig:dgcnn}
\end{figure}

Similar to point cloud constructed graph, patient's scan constructed graph edges are based on physical distance between slices (only nodes corresponding to successive slices are connected) as well as node's vector represents bounding boxes features which are based on euclidean space. Utilizing EdgeConv module into a graph based CNN to classify patient's scan represented by a graph described in section \ref{step 2} whether it contains an abnormal Adrenal. Part of a comprehensive library for Graph Deep Learning Yukuo Cen et al. \cite{cen2023cogdl} introduce DGCNN python implementation for graph classification tasks which used in this paper, resulting in NN architecture described in figure \ref{fig:dgcnn}. The first 2 layers are EdgeConv layers introduced by Wang et al. \cite{wang2019dynamic} then the output of those are concatenated (denote with $\oplus$) to fully connected layer (FC) with global max pooling operation, then multi-layer  perception (MLP) for achieving final classification scores.

\subsubsection{Step 4 - Detections Aggregation for localization} \label{step 4}

After previous step \ref{step 3} a class score of patient having an abnormal Adrenal is computed, the score value is in range $0$--$1$. A high score means that there is high probability for an abnormal Adrenal in whole patient's scan, finding the potential slices and the area defined by bounding boxes is important for a complete diagnosis. The following method approximates both for left and right Adrenals a slice together with it's bounding box withing the slices that have abnormal adrenal detections obtained across $k$ YOLO detectors shown in section \ref{step 2}, this approximated slice potentially have highest chance of having and abnormal Adrenal. 

Assume the following method preformed both for left and right Adrenals independently. Let patient's scan contains $S$ consecutive slices passed step "slices of interest" defined in section \ref{step 1}, let YOLO model $j$ $\forall j \in \{1..k\}$ for slice $i$ predicts score $score_{ij}$ - define the average score across $k$ YOLO models for single slice index $i$ as following $\overline{score}_{i}=\frac{\sum_{j=1}^{k} score_{ij}}{k} $ $\forall i \in \{1..S\}$. for scan index $i$ define score moving average with window size equals 5 as $ sma_i = \frac{\sum_{n=i-2}^{i+3} \overline{score}_{n}}{5} $, finally best scored slice index define  as $\hat{i} = \underset{i}{\mathrm{argmax}} (sma_i)$, note along with final slice index $\hat{i}$ we provide $r$ surrounding slices from each side. Let bounding box for  $slice_i$, $Yolo_j$ defined as $bb_{ij} \in \mathcal{R}^4$, final bounding box approximation defined as the average of all bounding boxes found in 5 surrounding slices of slice $\hat{i}$: $\hat{bbi} = \frac{\sum_{i=\hat{i}-2}^{\hat{i}+3} \sum_{j=1}^{k} bb_{ij}}{\lVert\{ bb_{ij}\}\lVert}$.

\subsection{Transfer Learning}
Insufficient training data is one of the most serious problems in deep learning, mostly in the clinical domain, where it is complex, expensive, and difficult to build a large-scale, high-quality annotated data-set. Transfer learning is an essential tool in machine learning to solve the basic problem of insufficient training data~\cite{tan2018survey}.

Tremendous progress has been made in image recognition, primarily due to the availability of large-scale annotated datasets such as the known ImageNet dataset (very comprehensive database of more than 1.2 million categorized natural images of 1000+ classes) and the recent revival of deep convolutional neural network. However, there exists no large-scale annotated medical image dataset comparable to ImageNet, as data acquisition is difficult, and quality annotation is costly~\cite{shin2016deep}. one of common techniques to  successfully employ CNNs to medical image classification is transfer learning, i.e., fine-tuning models pre-trained from natural image dataset to medical image tasks, or in other words instead of training CNNs from scratch using medical dataset with random initialization, the CNN trained on a known pre-trained model, which has millions natural well-labeled images in many classes, is fine-tuned to achieve the final task in the medical application such as classification or object detection. 

This research utilizes transfer learning in step "slices of interest" \ref{step 1} by reusing VGG-16 that was originally trained on a subset of the Image-Net dataset \cite{simonyan2014very}, also in step "abnormal adrenal detection" \ref{step 2} by reusing pretrained YOLO V3 model that was originally trained on Image-Net dataset as well. \cite{redmon2018yolov3}.

\section{ EVALUATION METHODS}
\label{scn:Eval}
This section introduces our evaluation methods for the 2 final tasks, classification and localization tasks in addition to intermediate steps of our MMGA method - 'slice of interest' and 'multi YOLO Abnormal Adrenal detections' steps.
% \Carmel{take ideas from section Evaluation Criteria in Object detection Survey paper}

\subsection{Experiment description}
\label{Experiment desc}

In this section, we present some statistics about our dataset and experimental description for our new method MMGA for Abnormal Adrenal detection on spinal CT scan. The dataset provided by the Soroka Medical Center from the orthopedics department. We provided Radiologists a third party application for labeling called \href{https://www.cvat.ai/}{cvat.ai}, the labeling process included looking at each slice for each scan and marking 'blue' bounding box around normal Adrenal and 'red' bounding box around abnormal Adrenal. Some statistics on the labeled dataset:

\begin{figure}[!h]
\centering
\includegraphics[scale=0.45]{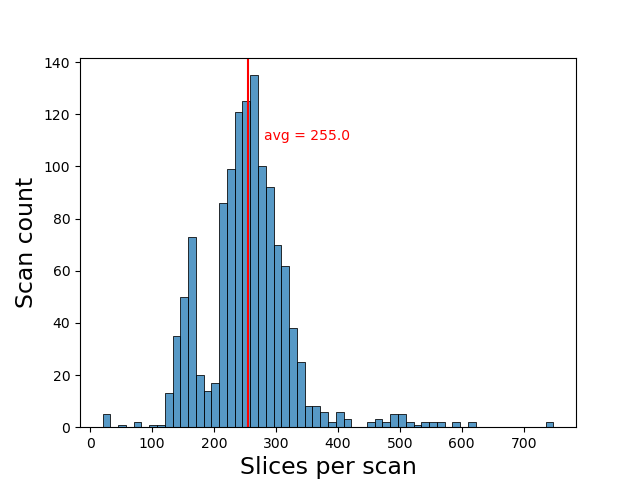}
\caption{Histogram showing number of slices in each scan} 
\label{fig:slice dist}
\end{figure}

\begin{figure}[h]
\begin{tabular}{ll}
\includegraphics[scale=0.48]{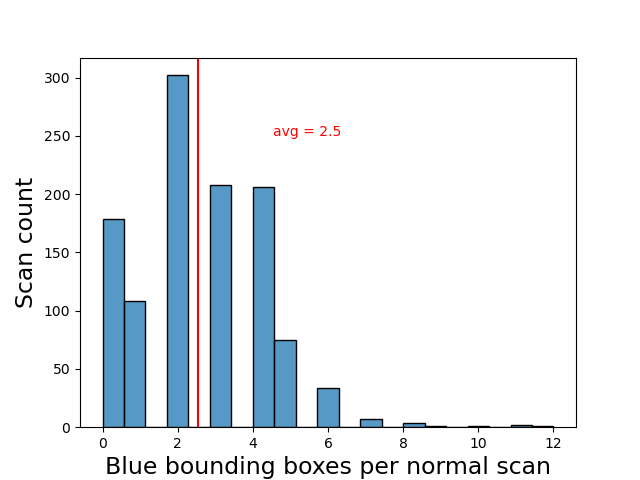}
&
\includegraphics[scale=0.48]{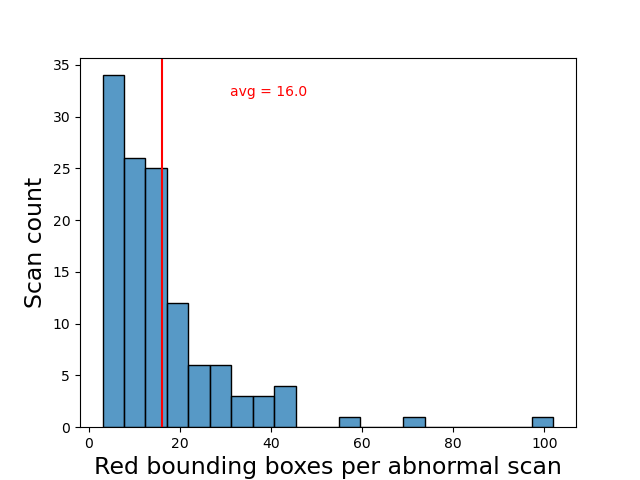}
\end{tabular}
\caption{Histogram showing bounding boxes count in each scan. Left: Normal scans, Right: Abnormal scans}
\label{fig:bb dist}
\end{figure}

\begin{enumerate}
\item The dataset includes 1250 scans (each scan corresponding to a patient), each scan includes 255 slices on average as shown in Figure \ref{fig:slice dist} along with histogram of slices per single scan.

% \begin{figure}[!h]
% \centering
% \includegraphics[scale=0.45]{Pictures/slice_dist.png}
% \caption{Histogram showing number of slices in each scan} 
% \label{fig:slice dist}
% \end{figure}

\item There 122 patients with at least 1 Abnormal adrenal (left or right) and 1128 patients with Normal Adrenals, in total 9.76\% from population has abnormal adrenal. 

\item\label{bb on avg} The data set includes 2.5 bounding boxes on average per scan for normal Adrenals across normal patent's scans as shown in Figure \ref{fig:bb dist}. not all slices within a scan with normal Adrenals are labeled only few selected slices from each patient's scan are labeled due to very time consuming labeling operation.

% \begin{figure}[h]
% \begin{tabular}{ll}
% \includegraphics[scale=0.48]{Pictures/Blue_bb_dist.png}
% &
% \includegraphics[scale=0.48]{Pictures/Red_bb_dist.png}
% \end{tabular}
% \caption{Histogram showing bounding boxes count in each scan. Left: Normal scans, Right: Abnormal scans}
% \label{fig:bb dist}
% \end{figure}

\item The data set includes 16 bounding boxes on average per scan for abnormal Adrenals across abnormal patent's scans as shown in Figure \ref{fig:bb dist}, all slices within each scan with abnormal Adrenals are labeled that is the ground truth for the final goal.

% \begin{figure}[!h]
% \centering
% \includegraphics[scale=0.9]{Pictures/Red_bb_dist.png}
% \caption{Histogram showing abnormal Adrenal bounding boxes in each abnormal scan} 
% \label{fig:red bb dist}
% \end{figure}

\item We split the dataset patient-wise having 30\% in test set and 70\% for training - 375 and 875 respectively. Within test-set 34 patients have abnormal Adrenals and 341 are normal scans. Within train-set 88 patients have abnormal Adrenals and 787 are normal scans.

\end{enumerate}

\begin{figure}[h]
\begin{tabular}{llll}
\includegraphics[scale=0.20]{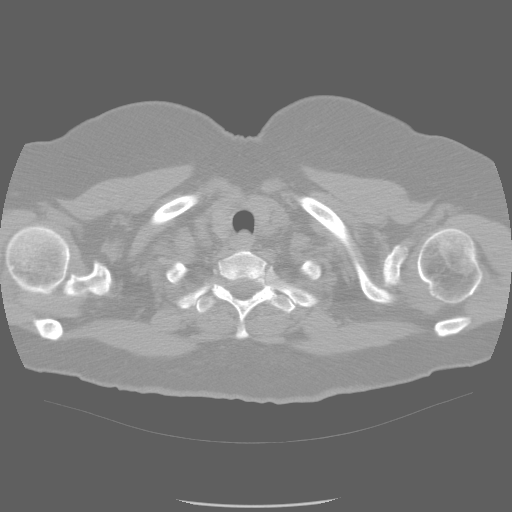}
&
\includegraphics[scale=0.20]{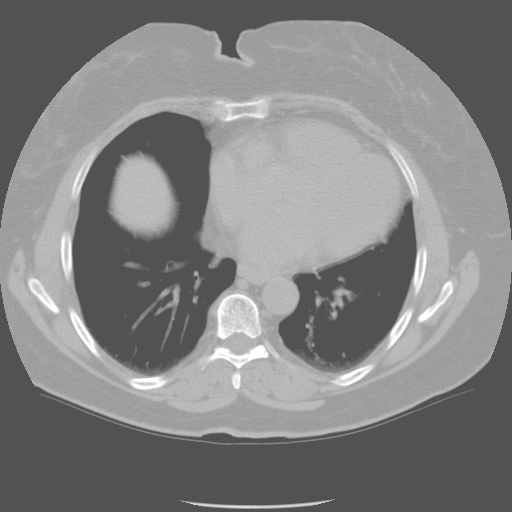}
&
\includegraphics[scale=0.20]{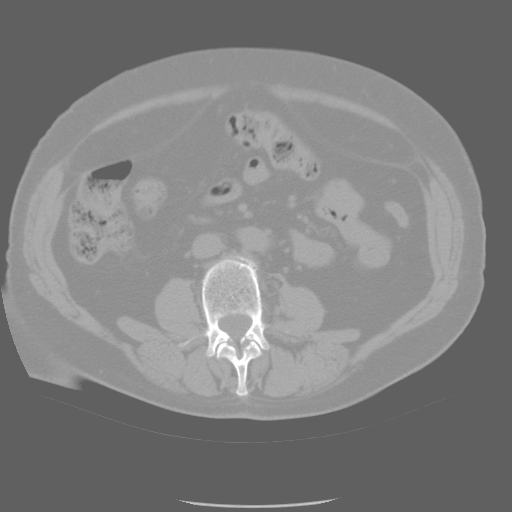}
&
\includegraphics[scale=0.20]{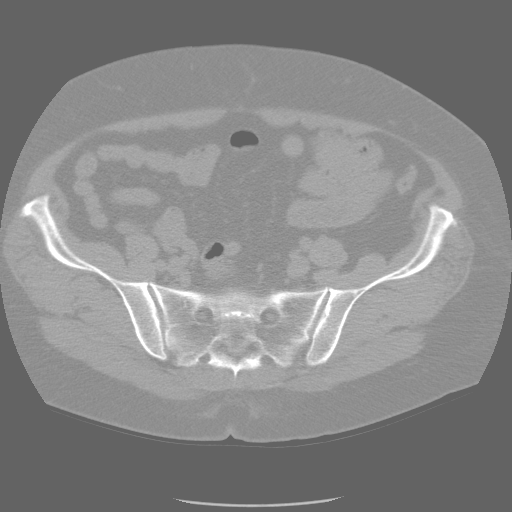}
\end{tabular}
\caption{Single scan slices at different axial (z) axis heights. From Left to right: Neck, Lungs, Abdomen, Sacrum.}
\label{fig:slices png}
\end{figure}

\begin{figure}[h]
\begin{tabular}{lll}
\includegraphics[scale=0.24]{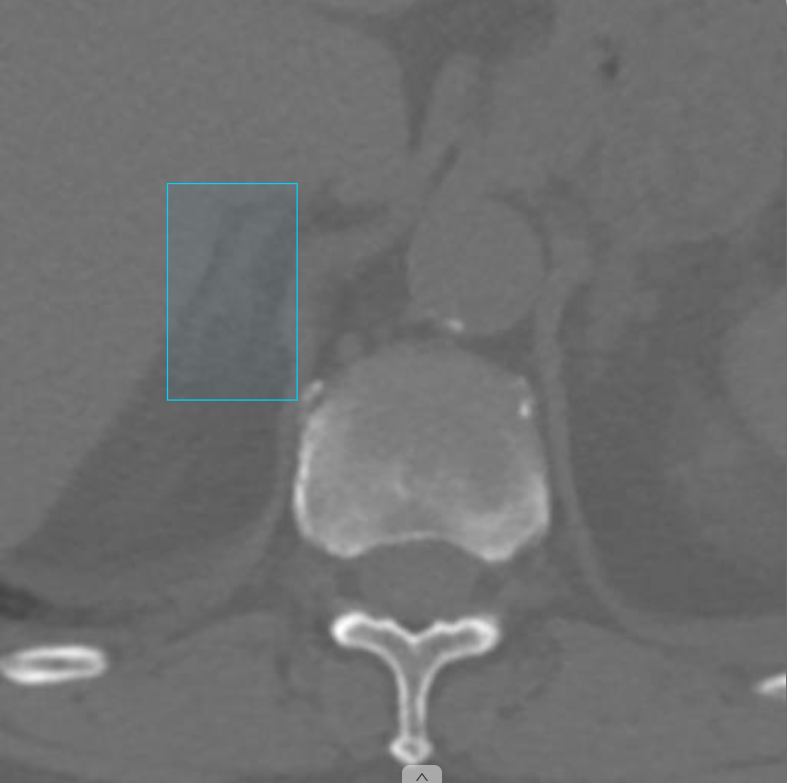}
&
\includegraphics[scale=0.24]{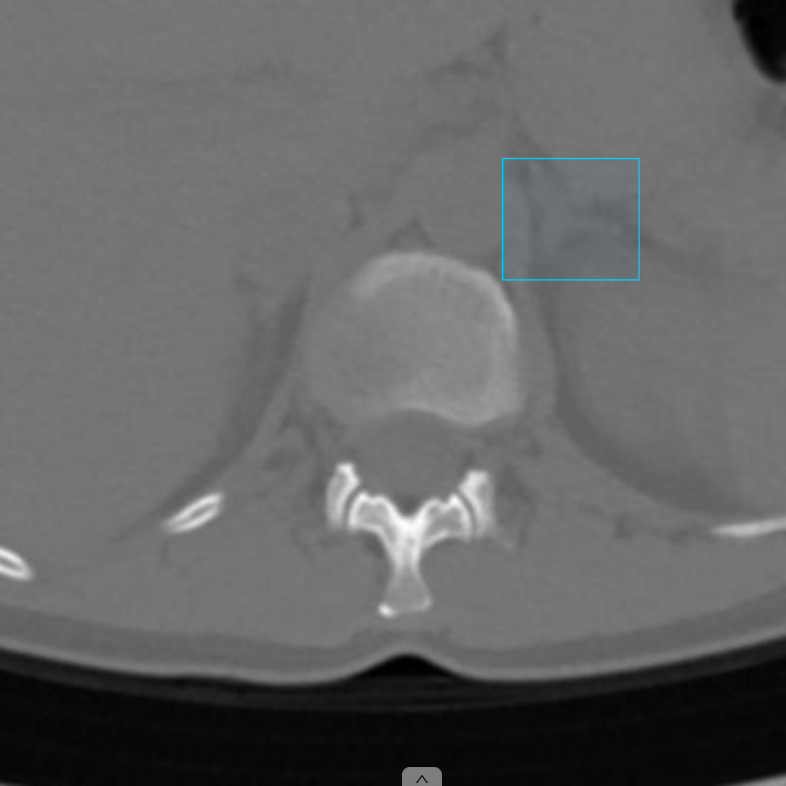}
&
\includegraphics[scale=0.24]{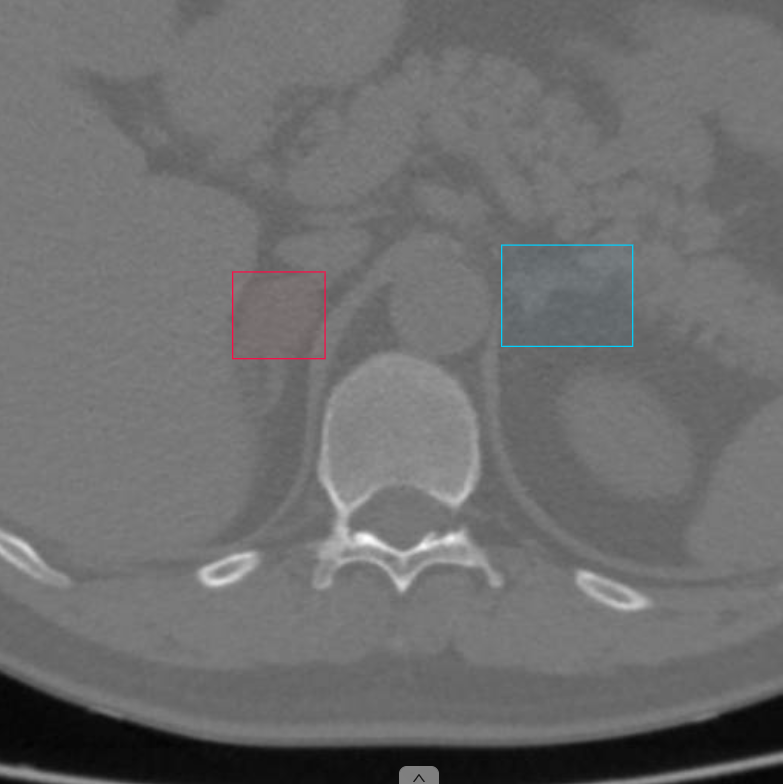}
\end{tabular}
\caption{Slices with bounding boxes, blue/red for normal/abnormal Adrenal respectively.}
\label{fig:adrenal slices}
\end{figure}

Figure~ \ref{fig:slices png} shows examples of slices at different height in Axial axis from a single patient scan, we can see major references in image's features between each and every image correspond to different Axial height. Figure~ \ref{fig:adrenal slices} shows examples of slices with red/blue (ab/normal) bounding boxes.
In experiments we use Radiologist's labeling as the ground truth for entire scan classification and for abnormal Adrenal localization. we evaluate model's goodness with respect to Radiologist's diagnoses, eventually this can be used as a second independent opinion for Radiologists.

% \begin{figure}[h]
% \begin{tabular}{llll}
% \includegraphics[scale=0.20]{Pictures/neck.png.png}
% &
% \includegraphics[scale=0.20]{Pictures/lungs.png.png}
% &
% \includegraphics[scale=0.20]{Pictures/abdomin.png.png}
% &
% \includegraphics[scale=0.20]{Pictures/sacrum.png.png}
% \end{tabular}
% \caption{Single scan slices at different axial (z) axis heights. From Left to right: Neck, Lungs, Abdomen, Sacrum.}
% \label{fig:slices png}
% \end{figure}

% \begin{figure}[h]
% \begin{tabular}{lll}
% \includegraphics[scale=0.24]{Pictures/blue_left_box.png}
% &
% \includegraphics[scale=0.24]{Pictures/blue_right_box.png}
% &
% \includegraphics[scale=0.24]{Pictures/red_blue_box.png}
% \end{tabular}
% \caption{Slices with bounding boxes, blue and red for normal abnormal Adrenal respectively}
% \label{fig:adrenal slices}
% \end{figure}

\subsection{Slices of interest measuring \ref{step 1}}
\label{SOI meas}

This step is important for reducing image variability, we built a CNN model predicting each slice whether it contains adrenal or not. The ground truth defined within each scan - all slices between first and last slice that have any adrenal label considered as "slices of interest" (positives) rest slices considered as negatives, some extra adjacent slices to each side (bottom/top) of the segment considered as slices of interest as well, that's because of the lack of normal Adrenal labels in every scan described in section \ref{Experiment desc} item \ref{bb on avg}, we use a validation set of 20\% from the initial test set. 

We evaluate the segment/slices of interest using 1 dimensional $IOU$ method as follow - let actual  slices of interest segment indices in a single scan defined as $Segment_{actual}=\{k, k+1,\dots\ , l \}$ where $k, k+1\dots\ , l$ are successive slice indices where Adrenals are found, similarly define for predicted slices of interest segment as 
$Segment_{pred}=\{p, p+1,\dots\ , q \}$, then define $IOU= \frac{|Segment_{pred} \bigcap Segment_{actual}|}{|Segment_{pred} \bigcup Segment_{actual}|}$, this metric evaluates predicted slices of interest segment v.s. actual segment for each scan. 

We defined another method to evaluate how well the predicted slices of interest segment \textbf{overlap} true segment.
Assume a scan $S$ contain $n$ slices we define $S=\{0, 1,\dots\ n-1 \}$ where $S \supseteq Segment_{actual}$ and $S \supseteq Segment_{pred}$, also define $Left_{\Delta} = \frac{p-k}{n}$ and $Right_{\Delta} = \frac{l-q}{n}$. Intuitively a good overlap would be positive close to zero value for $Left_{\Delta}$ and $Right_{\Delta}$, a negative value for $Left/Right_{\Delta}$ means that the predicted slices of interest fall short from true slices of interest.

\subsection{Multi YOLO Abnormal Adrenal detections}

At this intermediate step we trained 5 YOLO model each on 16K images (within slices of interest) without abnormal Adrenals (negatives) and 16K images (with repetitions) with abnormal Adrenals (positives), 10\% of images (positives and negatives independently) put aside for validation during training the models.
Each YOLO model started training from same weights position (YOLO-v3 \cite{redmon2018yolov3} that was trained on Darknet-53 datasets), each model was given partially different negative samples with random shuffle and same positive samples (with repetitions) with random shuffle as ground truth. 

we will evaluate this intermediate step for \textbf{slice} and \textbf{scan} level classifications using confusion matrix that will be introduced in next section (Figure~ \ref{fig:Confusion matrix} ), we define a simple aggregation methods for aggregating the multiple YOLO detections such as union and intersection. 

\begin{enumerate}
\item 
\textbf{\textit{Slice level evaluation}} - per each slice let each model's detection formulation as $Yolo_{i} \in \{0,1\}$. Model's positive detection of an abnormal adrenal in the image define as 1 and 0 when model did not detect abnormal adrenal. we define union aggregation of the multi models as $\bigcup_{i}Yolo_{i}=Yolo_{1} \lor \dots \lor Yolo_{n}$, similarly we define intersection aggregation as $\bigcap_{i}Yolo_{i}=Yolo_{1} \land \dots \land Yolo_{n}$. 
for $\bigcup_{i}Yolo_{i}$ a positive classification prediction for a slice/image is such that where at least one model predicts a bounding box (for abnormal adrenal) in the image.
The ground truth for the evaluation are slices with at least 1 abnormal adrenal bounding box taken as positive samples (615 in total, from 34 patient's scans), and all slices that have normal adrenal label and doesn't have any abnormal label are taken as negative samples (831 in total, from 289 patient's scans).
    
\item 
\textbf{\textit{Scan level evaluation}} - I use naive scan level classification method based on it's slices predictions, a scan per single YOLO model predicted as abnormal $\iff$ it contains at least 1 slice with abnormal adrenal bounding box.
The ground truth for the evaluation is all 375 scans from test-set (341 normal and 34 abnormal), filtering out from each scan all slices that are not in region of interest as described in step \ref{step 1}, in total there are 2264 slices of interest left from abnormal scans and 16075 slices left from normal scans.
A positive scan classification is such that where a model predicts a bounding box in any of patient's scan's slices, where as truth positive scan is such that where there is at least 1 abnormal Adrenal label in any of it's slices.

\end{enumerate}

% \textbf{\textit{Slice level evaluation}} - 
% Table~ \ref{tab:yolos conf mat} shows slice level classification confusion matrices for each of the 5 trained YOLO models, the ground truth for the evaluation are slices with at least 1 abnormal adrenal bounding box taken as positive samples (615 in total, from 34 patient's scans), and all slices that have normal adrenal label and doesn't have any abnormal label are taken as negative samples (831 in total, from 289 patient's scans). A successful model's slice/image classification is such that where a model predicts a bounding box (for abnormal adrenal) in the image.
% We can see each model solely perform poorly on slice level classification.

\subsection{Classification and Localization measuring}

After training 5 different YOLO models, each model aims to detect abnormal Adrenals (left and right) for every given slice (after initial slice filtering in step \ref{step 1}), we created a graph for each whole scan (described in setep \ref{step 3}) includes it's slices with their corresponding bounding boxes per each YOLO model. we trained graph learning DGCNN \cite{wang2019dynamic} based model for graph classification task, the two classes are 1 and 0 stand for ab/normal scan respectively.
In this reaserch CT scan pathology domain, the classification task is to categorize a CT scan into one of the following groups based on certain characteristics that the obtained model learned - 'Abnormal Adrenal found' or 'Normal Adrenal found'. 

\begin{figure}[!h]
\centering
\includegraphics[scale=0.6]{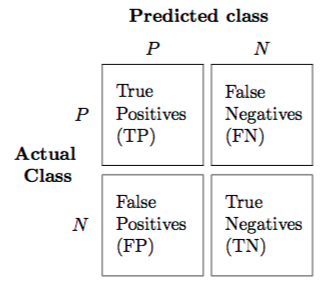}
\caption{Confusion matrix as a evaluation metric} 
\label{fig:Confusion matrix}
\end{figure}

Basically there are four possible outcomes of a classifier and a scan instance corresponding to the real value of the instance and the classifier's predicted value, as shown in Fig.\ref{fig:Confusion matrix}. This confusion matrix is the basis for many common metrics we use for evaluation:

\begin{enumerate}

    \item \textbf{\textit{Precision}} or PPV is the number of true positives (i.e. the number of items correctly labelled as belonging to the positive class) divided by the total number of elements labelled as belonging to the positive class (i.e. the sum of true positives and false positives):
    \begin{equation}
    Precision = \frac{TP}{TP + FP}
    \end{equation}
    
    \item \textbf{\textit{Recall}} also known as sensitivity is defined as the number of true positives divided by the total number of elements that actually belong to the positive class:
        \begin{equation}
    Recall = \frac{TP}{TP + FN}
    \end{equation}
    
    \item \textbf{\textit{F1-Score}} or F-Measurement is a measure of a test's accuracy, more precisely is the harmonic mean of precision and recall. The F-Measurement value ranges from zero to one, high F-Measurement values imply high classification efficiency. The following is the Formulation definition of F1-Score:
    \begin{equation}
    F1 = \frac{TP}{TP + 0.5(FP + FN)}
    \end{equation}
    
    \item \textbf{\textit{Receiver Operating Characteristic Curve}} (ROC) - is a two-dimensional graph shown in Figure~ \ref{fig:ROC}, the X axis is the FP rate and the Y axis in ROC graph is the TP rate.
    ROC are frequently used to show in a graphical way the connection/trade-off between advantages and costs for every possible cut-off for a test or a combination of tests, and  selecting better classifiers based on their performance.
    A ROC curve characteristics will determine what point is the best in terms of FN vs FP trade off within a classifier.
 
    \begin{figure}[!ht]
    \centering
    \includegraphics[scale=0.22]{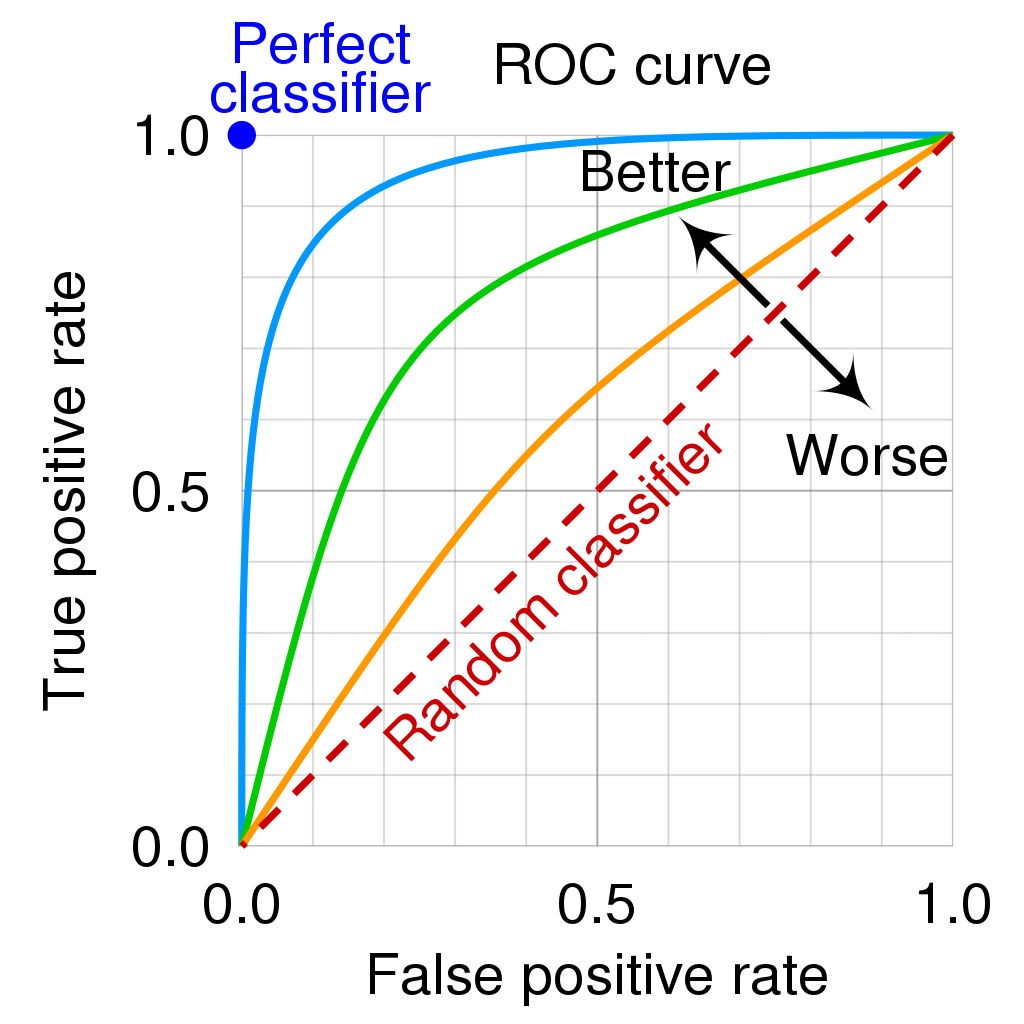}
    \caption{The ROC space for a "better" and "worse" classifier} 
    \label{fig:ROC}
    \end{figure}

    \item \textbf{\textit{Area Under a ROC}} Curve (AUC) -
    The area under the AUC illustrates the effectiveness of the model, The higher the AUC score the better is the model. When comparing between different classifiers an AUC is a single numeric value which is calculated as the area measured under ROC representing expected performance of a classifier. This is conventional and convenient way to compare classifiers.
    Furthermore the AUC of a classifier represents the probability that the classifier will rank a randomly selected positive instance higher than the randomly selected negative instance. 
    % Higher AUC score means better performance.
    
    % \begin{figure}[!h]
    % \centering
    % \includegraphics[scale=0.6]{Pictures/AUC.png}
    % \caption{AUC - ROC Curve} 
    % \label{fig:AUC}
    % \end{figure}
 
\end{enumerate}

% \subsection{Final Localization measuring}

\begin{figure}[!h]
    \centering
    \includegraphics[scale=1.1]{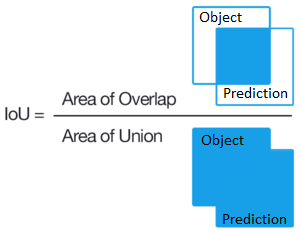}
    \caption{Computing the Intersection over Union is as simple as dividing the area of overlap between the bounding boxes by the area of union} 
    \label{fig:IoU}
    \end{figure}

After scan score achieved for abnormal Adrenal (positive) at step \ref{step 3} and based on a threshold cut-off value, we approximate the predicted slice index with highest chance to contain an abnormal Adrenal together with the bounding box approximation within that slice as described in Section \ref{step 3}. we evaluate the predicted slice index with \textbf{\textit{distance to closest slice to contain an abnormal Adrenal}} label (perfect score is 0 distance), while evaluating the bounding box is done with IOU method (explained below Figure~ \ref{fig:IoU}) between the true closest slice's bounding box to predicted slice's bounding box. Note - we do evaluation for left and right Adrenals independently. The basis for the evaluations are the TP slice only.

In 2D image object localization result often includes 4 numbers indicating object's bounding box description. The bounding box vector usually indicating 2 points from 4 points of a bounding box - top left (tl) and button right (br) $<X_{tl}, Y_{tl}, X_{br}, Y_{br}>$. The task is to localize the abnormal Adrenals (left and right) as close as possible to the ground truth bounding box, The following common method evaluate predicted bounding box v.s. true box. 
The \textbf{\textit{Intersection-Over-Union}} (IoU), also known as the \textit{Jaccard Index}, is one of the most commonly used metrics in semantic segmentation and detection tasks. IoU is an evaluation metric used to measure the accuracy of an object detector on a particular dataset. We often see this evaluation metric used in object detection challenges.
    
    % \item \textbf{\textit{Dice Coefficient}} (F1 Score) - is very similar to the IoU. They are positively correlated. Like the IoU, they both range from 0 to 1, with 1 signifying the perfect similarity between predicted and truth.
    % \begin{figure}[!h]
    % \centering
    % \includegraphics[scale=0.6]{Pictures/Dice.png}
    % \caption{the Dice Coefficient is 2 * the Area of Overlap divided by the total number of pixels in both images} 
    % \end{figure}

\section{RESULTS}
\label{scn:Experiments}

\subsection{Evaluating step "Slices Of Interest" \ref{step 1}}

\begin{figure}[!h]
\centering
\includegraphics[scale=0.48]{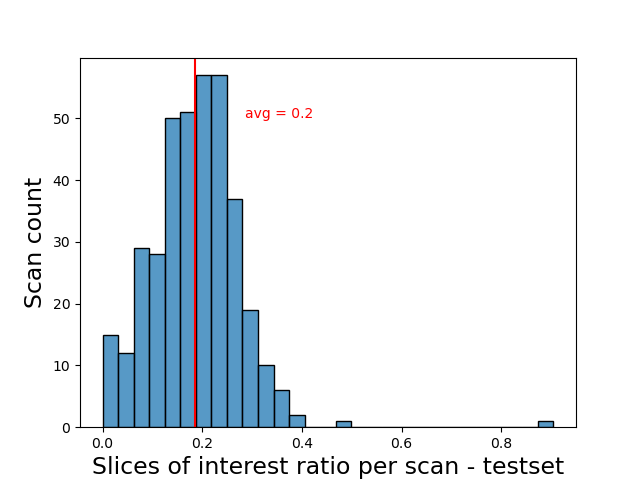}
\caption{Histogram showing ratio between predicted slices of interest to all slices per scan} 
\label{fig:left slice ratio}
\end{figure}

To illustrate how much images per scan were filtered out after 'slices of interest' step, Figure~ \ref{fig:left slice ratio} shows ratio between predicted slices of interest to all scan's slices, we can see on average around 80\% of slices are considered as not in interest narrowing to images around Adrenal space which is a substantial reduction of left image variability before passing to next step \ref{step 2}.

\begin{figure}[h]
\begin{tabular}{ll}
\includegraphics[scale=0.48]{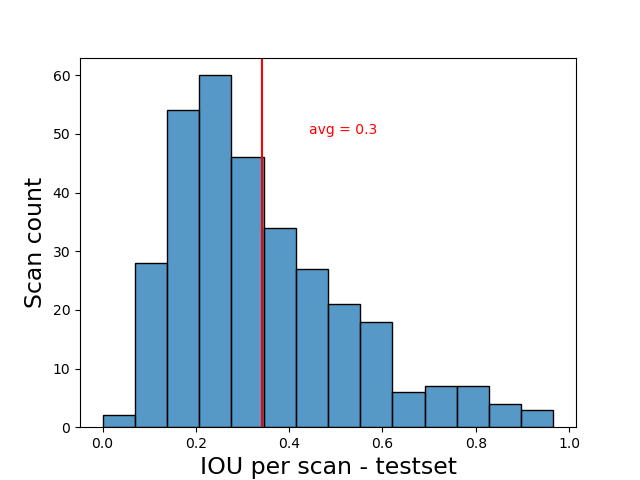}
&
\includegraphics[scale=0.48]{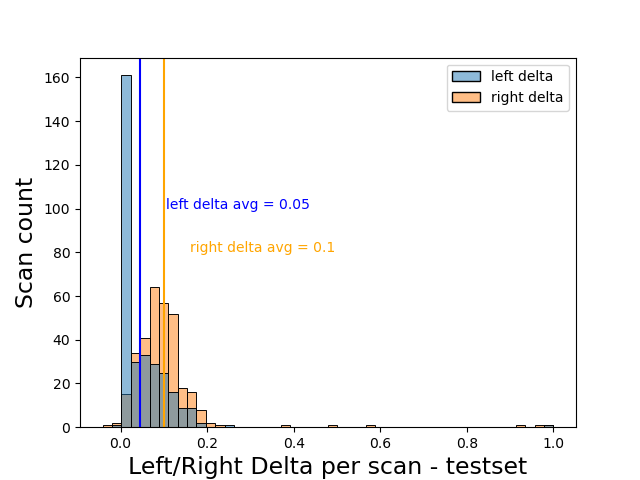}
\end{tabular}
\caption{Left: Slices of interest 1D IOU Histogram, Right: $Left_{\Delta}$ and $Right_{\Delta}$ histograms}
\label{fig:iou hist}
\end{figure}

% We performed 1 dimensional $IOU$ method as follow - let actual  slices of interest segment indices in a single scan defined as $Segment_{actual}=\{k, k+1,\dots\ , l \}$ where $k, k+1\dots\ , l$ are successive slice indices where Adrenals are found, similarly define for predicted slices of interest segment as 
% $Segment_{pred}=\{p, p+1,\dots\ , q \}$, then define $IOU= \frac{|Segment_{pred} \bigcap Segment_{actual}|}{|Segment_{pred} \bigcup Segment_{actual}|}$
Figure~ \ref{fig:iou hist} left plot shows IOU per scan summary in a histogram,  this metric evaluate predicted slices of interest segment v.s. true Adrenals segment for each scan. We can see that for some scans IOU value is very high (close to 1) and for some is very low, on average IOU is about 30\%.
Figure~ \ref{fig:iou hist} right plot shows $Left_{\Delta}$ and $Right_{\Delta}$ (defined in section \ref{SOI meas}) distributions we can see that the right boundary of detected segment overlaps true right boundary on average by 10\%  from total amount of slices in the scan, similarly the left boundary of detected segment overlaps true left bounder on average by 5\%, this deference is due to nature of given scans where Adrenal is seen right at the beginning of the scan (i.e. first slices, an illustration for such scan shown in Figure~ \ref{fig:slices proba}).
We can see very negligible negative values hence we conclude that predicted covers entirely actual slices of interest almost every scan. From both plots (Figure~ \ref{fig:iou hist}) we conclude that predicted slices of interest segment overlaps true Adrenal segment and on average the size of predicted segment is roughly 3 times the size of true segment since $IOU= \frac{|Segment_{pred} \bigcap Segment_{actual}|}{|Segment_{pred} \bigcup Segment_{actual}|} \overbrace{\sim}^{overlap} \frac{|Segment_{actual}|}{|Segment_{pred}|} \overbrace{\sim}^{avg} 0.3 \sim 1/3$.

% \begin{figure}[!h]
% \centering
% \includegraphics[scale=0.6]{Pictures/test_iou_slices_of_interest.png}
% \caption{Slices of interest IOU Histogram} 
% \label{fig:iou hist}
% \end{figure}

% \Carmel{here show how much on average slices in percent are left after applying this step show histogram (X: \% slices left from original scan) + can preform 1D IOU for evaluation}

\subsection{Evaluation Step "Abnormal Adrenal detections" \ref{step 2}}

% We Trained 5 YOLO model each on 16K images (within slices of interest) without abnormal Adrenals (negatives) and 16K images (with repetitions) with abnormal Adrenals (positives), 10\% of images (positives and negatives independently) put aside for validation.
% Each YOLO model started training from same weights position (YOLO-v3 \cite{redmon2018yolov3} that was trained on Darknet-53 datasets), each model was given partially different negative samples with random shuffle and same positive samples (with repetitions) with random shuffle as ground truth. 

\begin{figure}[h]
\begin{tabular}{llll}
\includegraphics[scale=0.20]{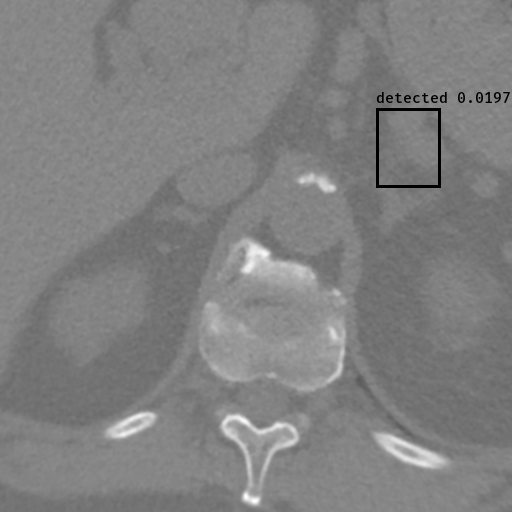}
&
\includegraphics[scale=0.20]{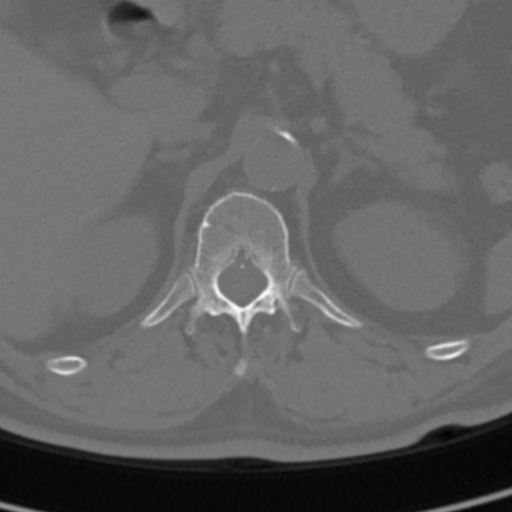}
&
\includegraphics[scale=0.20]{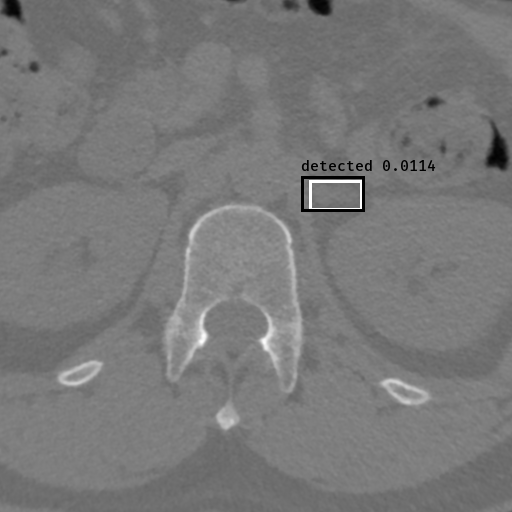}
&
\includegraphics[scale=0.20]{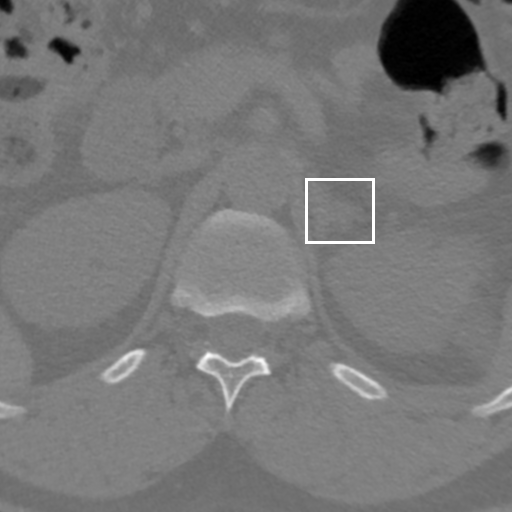}
\end{tabular}
\caption{Single YOLO predictions examples, from Left to right: FP, TN, TP, FN.}
\label{fig:all bb types}
\end{figure}

Figure~ \ref{fig:all bb types} show single YOLO model's image detection examples of slices with ab/normal adrenals, the white bounding boxes represent ground truth for abnormal Adrenals while the black bounding boxes represent predicted boxes and the black value represents the confidence score provided by YOLO model for predicted box to contain abnormal Adrenal.

\textbf{\textit{Slice level evaluation}} - 
Table~ \ref{tab:yolos conf mat} shows slice level classification confusion matrices for each of the 5 trained YOLO models, 
we can see each model solely perform poorly on slice level classification.
Table~ \ref{tab:pos slices pred} shows example for a typical abnormal patient's scan where there are several slices with abnormal adrenal, specifically shows the 5 trained YOLO models predictions only for 7 successive slices that contain abnormal adrenal, $1$ stands for successfully detected abnormal adrenal by a bounding box in single slice by a single model and $O$ for missing detecting the abnormal adrenal. We can see that different models detect the abnormal adrenal in different slices and some models (like $Yolo_2$) miss in every slice. Furthermore the multiple models disagree massively in their slice level detection as illustrated in Table~ \ref{tab:pos slices pred}. However a simple slice level union operation (last column)  $\bigcup_{i}Yolo_{i}$ show that all models complement each other for a wide range of slices with abnormal adrenals - this show the intuition behind the need for a multi model aggregation method explained in section \ref{step 3}.

\begin{table}[!h]
    \centering
\begin{tabular}{ |c||c|c|c|c|c|c| }
\hline
 %    &  \multicolumn{6}{|c|}{models' slice level confusion matrices} \\ 
 % \hline
Metric & $Yolo_1$  & $Yolo_2$ & $Yolo_3$ & $Yolo_4$  & $Yolo_5$ & Oracle\\ 
\hline
TP  & $152$ & $207$ & $155$ & $145$ & $120$ & 615\\ 
FN & 463 & 408 & 460 & 470 & 495 & 0\\ 
TN  & 776 & 792 & 800 & 794 & 792 & 831\\ 
FP & 55 & 39 & 31 & 37 & 39 & 0\\ 
 \hline
\end{tabular}
    \caption{Models classification performance - slice level confusion matrices}
    \label{tab:yolos conf mat}
\end{table}

\begin{table}[!h]
    \centering
\begin{tabular}{ |c||c|c|c|c|c|c|c| }
\hline
%     &  \multicolumn{7}{|c|}{partial scan's positive slices box predictions} \\ 
%  \hline
slice \# & $Yolo_1$  & $Yolo_2$ & $Yolo_3$ & $Yolo_4$  & $Yolo_5$ & $\bigcap_{i}Yolo_{i}$ & $\bigcup_{i}Yolo_{i}$    \\ 
\hline
 1 & 1  & O & O & O & O & O & 1\\ 
 2 & O & O & O & O & O & O & O\\ 
 3 & O & O & O & O & 1  & O & 1\\ 
 4 & O & O & 1  & 1  & O & O & 1\\ 
 5 & 1  & O & O & O & O & O & 1\\ 
 6 & 1  & O & O & O & O & O & 1\\ 
 7 & O & O & O & 1  & O & O & 1\\ 
 \hline
\end{tabular}
    \caption{Example for partial scan's positive slices bounding box predictions}
    \label{tab:pos slices pred}
\end{table}

\textbf{\textit{Scan level evaluation}} - 
% We use naive scan level classification method based on it's slices predictions, a scan predicted as abnormal $\iff$ it contains at least 1 slice with abnormal adrenal bounding box.
% The ground truth for the evaluation is all 375 scans from test-set (341 normal and 34 abnormal), filtering out from each scan all slices that are not in region of interest as described in step \ref{step 1}, in total there are 2264 slices of interest left from abnormal scans and 16075 slices left from normal scans.
% A positive scan classification is such that where a model predicts a bounding box in any of patient's scan's slices, where as truth positive scan is such that where there is at least 1 abnormal Adrenal label in any of it's slices. 
Let $scans_{pos}=\{pos_1,\dots\ , pos_k \}$ define the set of positively labeled (ground truth) scans ids, similarly $cans_{neg}=\{neg_1,\dots\ , neg_l \}$ negatively labeled scan ids. For each model $Yolo_i$ for each metric $met \in \{TP, FN, TN, FP\}$ we define $Yolo_{i,met}=\{scan_{1, met},\dots\ , scan_{i, met}\}$ a set of all classified scans ids under metric $met$ by $Yolo_i$.
we define set $TP\bigcap_{i}Yolo_{i}=\{pos_j|pos_j \in \bigcap_{i}Yolo_{i,TP}\}$ and complementary define set $FN\bigcap_{i}Yolo_{i}=\{pos_j|pos_j \notin \bigcap_{i}Yolo_{i,TP}\}$, similarly we define set $FP\bigcap_{i}Yolo_{i}=\{neg_j|neg_j \in \bigcap_{i}Yolo_{i,FP}\}$ and complementary define set $TN\bigcap_{i}Yolo_{i}=\{neg_j|neg_j \notin \bigcap_{i}Yolo_{i,FP}\}$. In the same manner we define the all union sets $met\bigcup_{i}Yolo_{i}$ for each $met \in \{TP, FN, TN, FP\}$.

% Let each model for each metric $Yolo_{i,metric}=\{scan_1,\dots\ , scan_{i, met} \}$ denotes all model's classified patients/scans, we define for each metric the $\bigcap_{i}Yolo_{i}$ and $\bigcup_{i}Yolo_{i}$ sets for intersection and union sets respectively.
Table~ \ref{tab:best_clf_scores} shows a naive scan level classification evaluation for each of the 5 trained YOLO models together with two simple aggregation methods (union and intersection).
We can see the multiple models widely agree within abnormal patients on positively detected (28 out of 34) as shown in column $\bigcap_{i}Yolo_{i}$ under $TP$ metric. 
Each YOLO model has very high FP rate while there is less agreement between models (only 87 patient out of 341) within normal patients that falsely detected as abnormal is shown in column $\bigcap_{i}Yolo_{i}$ under $FP$ metric i.e. FP rate is relatively lower against each model solely.
$\bigcup_{i}Yolo_{i}$ FP value is very high showing that almost every normal scan (313/341)  has an abnormal Adrenal prediction in some portion of it's slices.
% this is another showing the intuition behind the need for a multi model aggregation method explained in section \ref{step 3}.

\begin{table}[!h]
    \centering
\begin{tabular}{ |c||c|c|c|c|c|c|c|c| }
\hline
 %    &  \multicolumn{7}{|c|}{Positively detected patients/scans} \\ 
 % \hline
Metric & $Yolo_1$  & $Yolo_2$ & $Yolo_3$ & $Yolo_4$  & $Yolo_5$ & $\bigcap_{i}Yolo_{i}$ & $\bigcup_{i}Yolo_{i}$  & Oracle \\ 
\hline
TP  & 32 & 30 & 31 & 33 & 31 & 28 & 33 & 34\\
FN  & 2 & 4 & 3 & 1 & 4 & 6 & 1 & 0\\
TN & 104 & 122 & 171 & 123 & 122 & 254 & 28 & 341\\ 
FP & 237 & 219 & 170 & 218 & 219 & 87 & 313 & 0\\ 
 \hline
\end{tabular}
    \caption{YOLO models performance naive patients classification}
    \label{tab:best_clf_scores}
\end{table}

\subsection{Final Scan Classification Evaluation}

\begin{figure}[!h]
\centering
\includegraphics[scale=0.39]{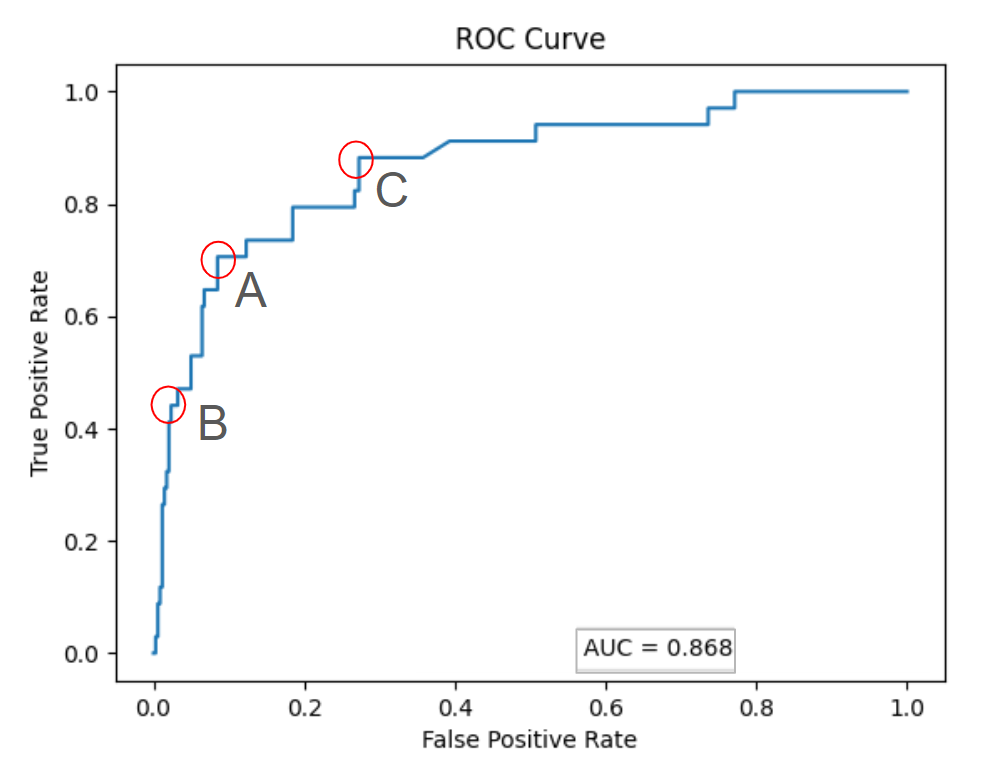}
\caption{ROC curve for our MMGA model} 
\label{fig:roc dgcnn}
\end{figure}

% \begin{table}[!h]
%     \centering
% \begin{tabular}{ |c||>{\columncolor[RGB]{107, 228, 119}}c|>{\columncolor[RGB]{255, 128, 0}}c|>{\columncolor[RGB]{255, 102, 178}}c|c| }
% \hline
%  %    &  \multicolumn{7}{|c|}{Positively detected patients/scans} \\ 
%  % \hline
% Metric & $B$  & $A$ & $C$ & Oracle\\ 
% \hline
% TP  & 14 (41.2\%)  & 24 (70.6\%) & 30 (88.2\%)& 34 \\
% FN  & 20  (58.8\%) & 10  (29.4\%) & 4 (11.8\%)& 0\\
% \hline
% TN  & 334 (98\%) & 312 (91.5\%)& 245 (71.8\%)& 341\\ 
% FP  & 7 (2\%)   & 29 (8.5\%) & 96 (28.2\%)& 0\\ 
%  \hline
% PPV & 0.667   & 0.453  & 0.24 & 1\\ 
% NPV & 0.944   & 0.969  & 0.984 & 1\\ 
% recall & 0.412   & 0.706  & 0.882 & 1\\
% F1 & 0.509   & 0.552  & 0.377 & 1\\
%  \hline
% \end{tabular}
%     \caption{different operational cut-off points scan classification performance evaluations}
%     \label{tab:a b c conf}
% \end{table}

\begin{table}[!h]
    \centering
\begin{tabular}{ |c||c|c|c|c| }
\hline
 %    &  \multicolumn{7}{|c|}{Positively detected patients/scans} \\ 
 % \hline
Metric & $B$  & $A$ & $C$ & Oracle\\ 
\hline
TP  & 14  & 24  & 30 & 34 \\
FN  & 20  & 10  & 4 & 0\\
TN  & 334 & 312 & 245 & 341\\ 
FP  & 7   & 29  & 96 & 0\\ 
 \hline
PPV/precision & 0.667   & 0.453  & 0.24 & 1\\ 
NPV & 0.944   & 0.969  & 0.984 & 1\\ 
recall & 0.412   & 0.706  & 0.882 & 1\\
F1 & 0.509   & 0.552  & 0.377 & 1\\
 \hline
\end{tabular}
    \caption{different operational cut-off points scan classification performance evaluations}
    \label{tab:a b c conf}
\end{table}

Figure~ \ref{fig:roc dgcnn} show ROC curve as an evaluation for our compound MMGA trained model with highest AUC score reaching 0.869 among other trained models within hyper parameter space (its hyper parameters: lr=0.01, hidden size=64, dropout=0.75, k=30, weight decay=0.05). we marked 3 operational points options on the ROC curve $A$, $B$ and $C$ that demonstrate different needs and aspects from user/business point of view. 

\begin{enumerate}
\item Point $A$ can be a good default operational choice when there are no requirements, this point is chosen based on distance to top left corner where the closest point on curve is chosen, the cut-off threshold value for point $A$ is 0.259.

\item Point $B$ can be a good choice when it's important for FPR to be low for example when the treatment cost of positive prediction is high, the cut-off threshold value for point $B$ is 0.506.

\item Point $C$ is a reasonable choice when it's very important not missing the anomaly for example when there is high risk of having the illness, the cut-off threshold value for point $C$ is 0.162.

\end{enumerate}

% \begin{table}[!h]
%     \centering
% \begin{tabular}{ |c||c|c|c|c| }
% \hline
%  %    &  \multicolumn{7}{|c|}{Positively detected patients/scans} \\ 
%  % \hline
% Metric & $B$  & $A$ & $C$ & Oracle\\ 
% \hline
% TP  & 14  & 24  & 30 & 34 \\
% FN  & 20  & 10  & 4 & 0\\
% TN  & 334 & 312 & 245 & 341\\ 
% FP  & 7   & 29  & 96 & 0\\ 
%  \hline
% PPV/precision & 0.667   & 0.453  & 0.24 & 1\\ 
% NPV & 0.944   & 0.969  & 0.984 & 1\\ 
% recall & 0.412   & 0.706  & 0.882 & 1\\
% F1 & 0.509   & 0.552  & 0.377 & 1\\
%  \hline
% \end{tabular}
%     \caption{different operational cut-off points scan classification performance evaluations}
%     \label{tab:a b c conf}
% \end{table}

The performance evaluation corresponding to operational points $A$, $B$ and $C$  is shown in Table \ref{tab:a b c conf}.   
Although the prior probability of a suspicious lesion in the adrenal in our data set is only 9.76\%, operational point $B$ shows we have reached a positive predictive value [PPV] of up to 67\% -- an almost 7-fold higher risk for a truly suspicious lesion, compared to the background prevalence.
On the other hand, although the prior probability of a patient NOT having an abnormality is 90.24\%, our negative predictive value [NPV] is up to 94.4\%, reducing the risk of an abnormality, when our system says that the adrenal seems normal, from 9.76\%  down to 5.6\%, i.e. a relative reduction of around 42\% of the risk.

% reaching PPV up to 67\% that is big improvement of odds time 6.8 (since the apriori abnormal patient rate in population is 9.76\%), while NPV get to 94.4\% so the risk of having abnormality is lowered from 9.76\% to 5.6\%.

% \begin{table}[!h]
%     \centering
% \begin{tabular}{ |c||c|c|c|c| }
% \hline
%  %    &  \multicolumn{7}{|c|}{Positively detected patients/scans} \\ 
%  % \hline
% Metric & $B$  & $A$ & $C$ & Oracle\\ 
% \hline
% TP  & 14  & 24  & 30 & 34 \\
% FN  & 20  & 10  & 4 & 0\\
% TN  & 334 & 312 & 245 & 341\\ 
% FP  & 7   & 29  & 96 & 0\\ 
%  \hline
% PPV/precision & 0.667   & 0.453  & 0.24 & 1\\ 
% NPV & 0.944   & 0.969  & 0.984 & 1\\ 
% recall & 0.412   & 0.706  & 0.882 & 1\\
% F1 & 0.509   & 0.552  & 0.377 & 1\\
%  \hline
% \end{tabular}
%     \caption{different operational cut-off points scan classification performance evaluations}
%     \label{tab:a b c conf}
% \end{table}

\textbf{Graph learning v.s. simple rule based classification method}, 
Our final step \ref{step 3} for classification is complex graph deep learning based method, we will test a simple majority vote rule based method against it. 
% A scan/patient will be predicted as abnormal if there will be $S$ successive slices such that each slice should be predicted as abnormal by majority voting of our 5 YOLO models.

% \subsubsection{Graph learning v.s. simple rule based classification method}
\label{experiments:graph vs simple}
\begin{table}[!h]
    \centering
\begin{tabular}{ |c||c|c|c|c|c|c|c|c|c|c|c|c| }
\hline
    &  \multicolumn{12}{|c|}{minimum successive positive slices by majority vote} \\ 
 \hline
Metric & 0  & 1 & 2 & 3 & 4  & 5 & 6 & 7 & 8 & 9 & 10 & 11\\ 
\hline
TP  & 34 & 26  & 19 & 11 & 6  & 4  & 2  & 2 & 1   & 1  & 1  &       0 \\
FN  & 0  & 8  & 15  & 23 & 28 & 30 & 32 & 32 & 33 & 33  & 33 &       34 \\
TN  & 0 & 257  & 304  & 325 & 333 & 337& 338 &   340& 340& 341  & 341 & 341 \\ 
FP  & 341  & 84  & 37 & 16 & 8    & 4  & 3 & 1 & 1 & 0  & 0 & 0 \\ 
 \hline
PPV & 0.09 & 0.24  & 0.34 & 0.41 & 0.43 & 0.5  & 0.4  & 0.67 & 0.5 & 1.0  & 1.0  & 0.0\\ 
NPV & 0.0  & 0.97  & 0.95 & 0.93 & 0.92 & 0.92 & 0.91 & 0.91 & 0.91& 0.91 & 0.91 & 0.91\\ 
recall&1.0 & 0.76  & 0.56 & 0.32 & 0.18 & 0.12 & 0.06 & 0.06 & 0.03& 0.03 & 0.03 & 0.0\\
F1 & 0.17  & 0.36  & 0.42 & 0.36 & 0.25 & 0.19 & 0.1  & 0.1  & 0.06& 0.06 & 0.06 & 0.0\\
 \hline
\end{tabular}
    \caption{different operational cut-off points based on successive positive slices by majority vote - classification performance evaluations}
    \label{tab:simple rule conf}
\end{table}

\begin{figure}[!h]
\centering
\includegraphics[scale=0.68]{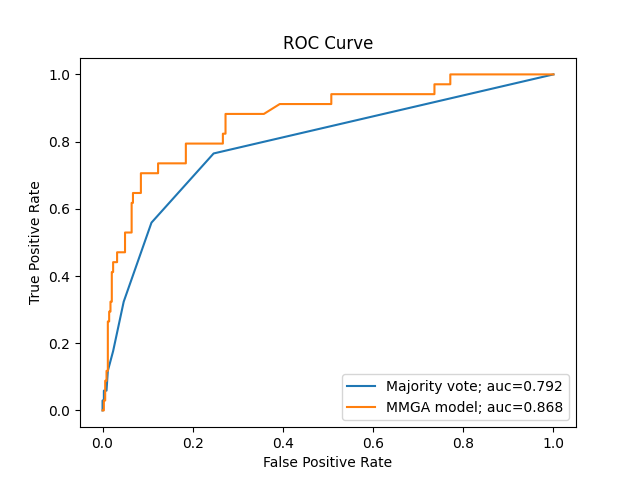}
\caption{ROC curve for a simple majority based method and proposed MMGA method} 
\label{fig:roc simple rule}
\end{figure}

A scan/patient will be predicted as abnormal if there will be $S$ successive slices such that each slice should be predicted as abnormal by majority voting of our 5 YOLO models.
Table \ref{tab:simple rule conf} show classification evaluation results, we can compare to operational point $A$ of our MMGA to this method where "minimum successive positive slices" equals to 2: This method's FP is 37 while the MMGA is better with 29 only furthermore MMGA reaching TP of 24 while this simple method's TP gets only to 19.

% \begin{table}[!h]
%     \centering
% \begin{tabular}{ |c||c|c|c|c|c|c|c|c|c|c|c|c| }
% \hline
%     &  \multicolumn{12}{|c|}{minimum successive positive slices by majority vote} \\ 
%  \hline
% Metric & 0  & 1 & 2 & 3 & 4  & 5 & 6 & 7 & 8 & 9 & 10 & 11\\ 
% \hline
% TP  & 34 & 26  & 19 & 11 & 6  & 4  & 2  & 2 & 1   & 1  & 1  &       0 \\
% FN  & 0  & 8  & 15  & 23 & 28 & 30 & 32 & 32 & 33 & 33  & 33 &       34 \\
% TN  & 0 & 257  & 304  & 325 & 333 & 337& 338 &   340& 340& 341  & 341 & 341 \\ 
% FP  & 341  & 84  & 37 & 16 & 8    & 4  & 3 & 1 & 1 & 0  & 0 & 0 \\ 
%  \hline
% PPV & 0.09 & 0.24  & 0.34 & 0.41 & 0.43 & 0.5  & 0.4  & 0.67 & 0.5 & 1.0  & 1.0  & 0.0\\ 
% NPV & 0.0  & 0.97  & 0.95 & 0.93 & 0.92 & 0.92 & 0.91 & 0.91 & 0.91& 0.91 & 0.91 & 0.91\\ 
% recall&1.0 & 0.76  & 0.56 & 0.32 & 0.18 & 0.12 & 0.06 & 0.06 & 0.03& 0.03 & 0.03 & 0.0\\
% F1 & 0.17  & 0.36  & 0.42 & 0.36 & 0.25 & 0.19 & 0.1  & 0.1  & 0.06& 0.06 & 0.06 & 0.0\\
%  \hline
% \end{tabular}
%     \caption{different operational cut-off points based on successive positive slices by majority vote - classification performance evaluations}
%     \label{tab:simple rule conf}
% \end{table}

Figure~ \ref{fig:roc simple rule} shows ROC curve for both the simple rule based method with AUC score of 0.792 and our proposed MMGA with AUC score of 0.868. Comparing the simple majority vote base method to our MMGA model, its clear that the ROC curve of MMGA is superior to the other simpler method in addition to much higher AUC value, thus concluding that there is benefit in applying another ML aggregation method on top of the Multi YOLO models detections to enhance classification Step, furthermore our proposed MMGA method better against the simple majority vote base method for scan classification. 

\subsection{Final Abnormal Adrenal Localization Evaluation}

% After abnormal Adrenal scan score achieved and based on threshold cut-off value for operational points $A$, $B$ and $C$, we approximate the predicted slice index with highest chance to contain an abnormal Adrenal together with the bounding box approximation within that slice as described in Section \ref{step 3}. we evaluate the predicted slice index with distance to closest slice to contain an abnormal Adrenal label (perfect score is 0 distance), while evaluating the bounding box is done with IOU method between the closest slice's bounding box to predicted slice's bounding box. Note - we do evaluation for left and right Adrenals independently. For each operational point the basis for the evaluations are the TP slice only.

Table \ref{tab:a b c conf} show the average evaluation scores for $A$, $B$ and $C$ operational points, we can see the evaluation scores for right Adrenal both for IOU and distance to abnormal slice is much better than for left Adrenal, that can be explained by the abundance of right abnormal Adrenal labels examples compare to left ones. In train-set we have 225:1055 labels for left and right abnormal Adrenal respectively, while in test-set a similar ratio of 109:564.
We can see a correlation between left IOU scores (left Adrenal are the majority of incidents) to operational points cut-off thresholds, the more strict/higher cut-off threshold (threshold value represent the minimum probability for predicting a scan as abnormal) the higher IOU average score - that make sense since abnormal MMGA's predicted lower probability implies higher uncertainty for a scan to have Abnormal Adrenal or in other words - its harder to detect it which can effect on poorer predicted bounding boxes. 

\begin{table}[!h]
    \centering
\begin{tabular}{ |c||c|c|c|c| }
\hline
 %    &  \multicolumn{7}{|c|}{Positively detected patients/scans} \\ 
 % \hline
Metric & $B$  & $A$ & $C$ & Oracle\\ 
\hline
cut-off th & 0.506  & 0.259  & 0.162 & NA \\
left avg IOU  & 0.395  &  0.364  & 0.408 & 1\\
right avg IOU & 0.602 & 0.593 & 0.522 & 1\\ 
left avg $\Delta$  & 8.2   & 9.1  & 8.2 & 0\\ 
right avg $\Delta$  & 1.3   & 1.0  & 2.25 & 0\\ 
 \hline
TP (eval base)  & 14  & 24  & 30 & 34 \\
\hline
\end{tabular}
    \caption{Abnormal Adrenal localization evaluation}
    \label{tab:loc eval}
\end{table}

\begin{figure}[h]
\begin{tabular}{ll}
\includegraphics[scale=0.23]{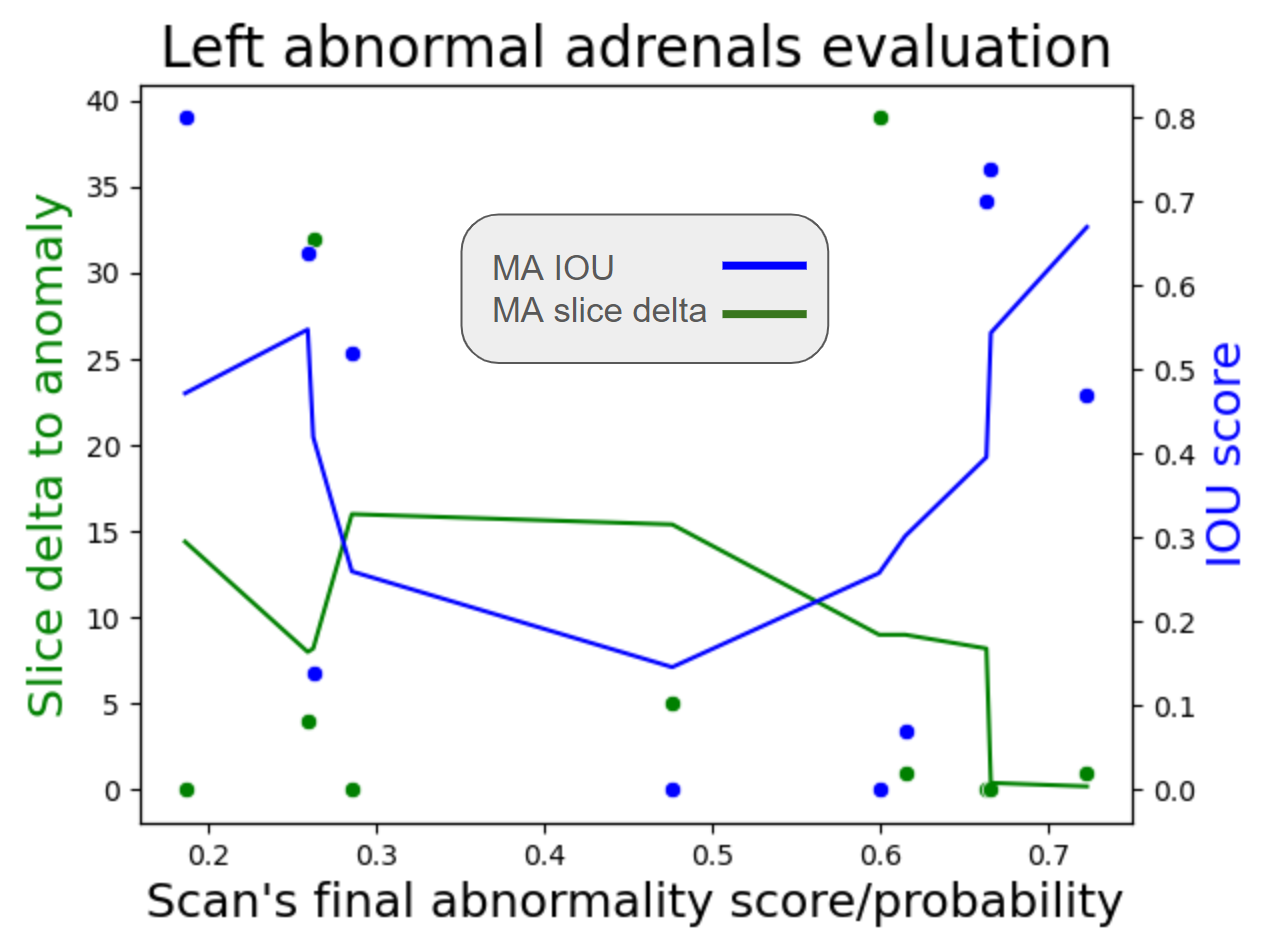}
&
\includegraphics[scale=0.23]{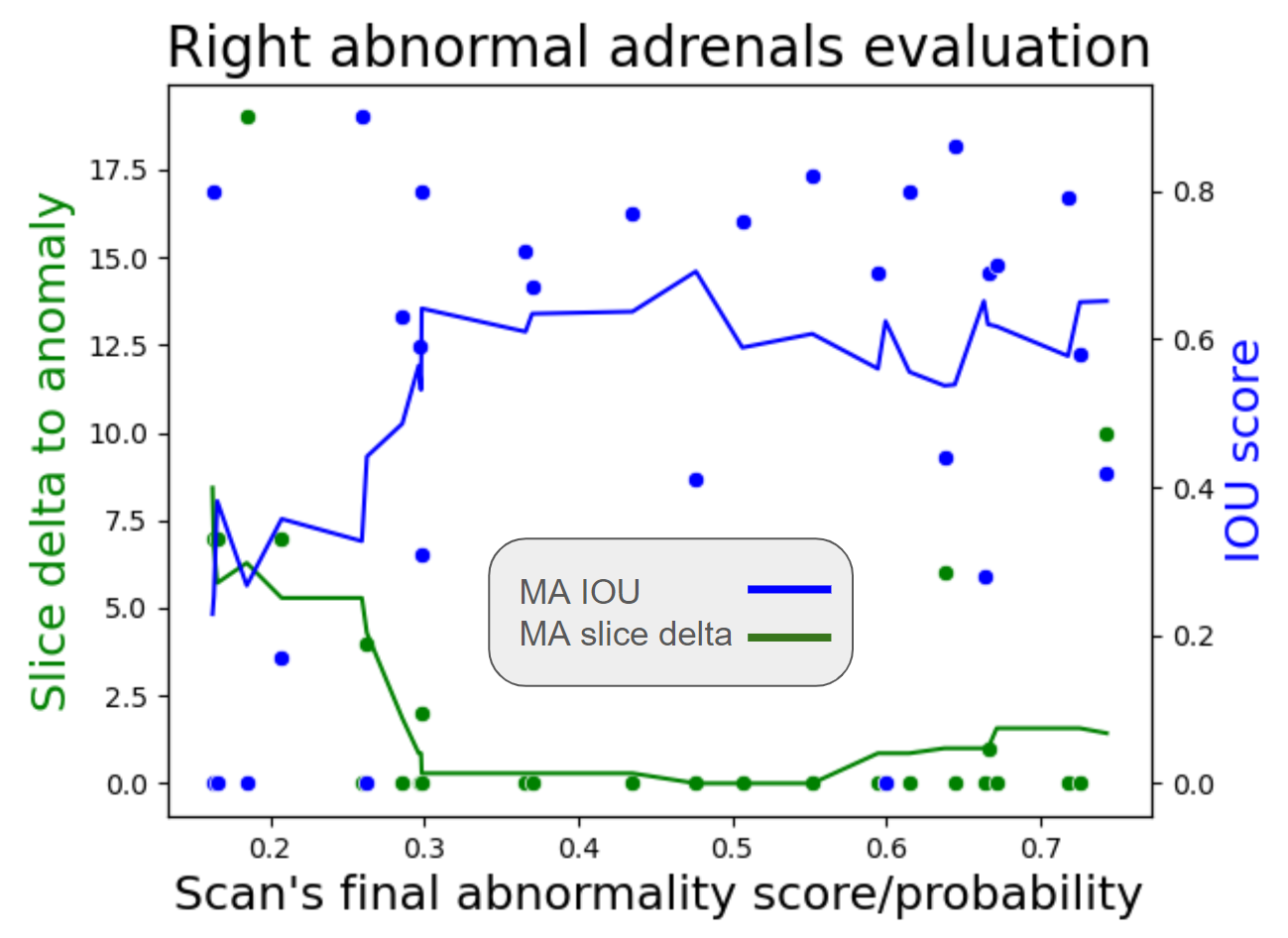}
\end{tabular}
\caption{Left / Right Abnormal Adrenals localization evaluation}
\label{fig:left right eval}
\end{figure}

Figure~  \ref{fig:left right eval} represent operational point $C$ localization evaluation which includes 30 TP scans, both for left and right Adrenals.
The $X$ axis represent final scan's abnormality score achieved by our MMGA model, left $Y$ axis represents the distance between predicted slice index to closest slice containing actual abnormal Adrenal and finally the right $Y$ axis represents IOU score between predicted bounding box to actual bounding box at the closest slice containing actual abnormal Adrenal, the lines in the plot represent moving average with window size of 5 and 7 for left and right respectively Adrenals.
we can see a very reasonable correlation between "slice delta to anomaly" to IOU score, i.e. when the IOU is higher/better the slice delta to anomaly is lower/better.

% \section{FUTURE WORK}
% \label{scn:FutureWork}
% \input{Future}

\section{SUMMARY AND DISCUSSION}
\label{scn:Summary}
\subsection{Main Contributions}

Developing an automatic system for flagging abdominal abnormalities such as malignancies or cancer in Adrenals that will eventually enters the radiologists' routine, will reduce the work load and improve diagnoses' accuracy or at least serve as another independent opinion that can strengthen radiologist's diagnosis. Radiologists will have more time and attention to focus on the spinal aspects within a scan since the patient originally complains about back issues.
Such system can alert on suspicious patients with anomaly within an organ and give pointers to specific images/slices with bounding boxes marking the suspicious anomaly within an image, which would drastically reduce the load work on radiologists and improve detection rate of anomalous organs.
Due to work load reduction patients will benefit a much shorter response time and the treatment will start much faster. This system can be used for analysing old CT scans and contribute to medical reaserch. 

Our novel method using only CT scans for identifying anomalous Adrenals, this method can be utilized to other types of scans like MRI and also detecting anomalies in other organs not restricted to abdomen. 

This compound method aggregates bounding boxes obtained from different slices from a single scan, similarly the architecture can be beneficial to other detection applications that involve multiple ordered unlimited amount of images connected to each other such as videos that contain multiple images as well.

\subsection{Advantages And Limitations}
Our compound method is not restricted by the number of slices to which the scan to be evaluated/trained has been partitioned; it uses the power of Graphs to create a single entity graph from all slices to be predicted with Graph Learning techniques.
Furthermore, our MMGA method is not restricted by the size of the images of a scan to be evaluated. For example, we trained the model on images of size 512x512 pixels each, but we can evaluate scans with different image sizes as well, by using  simple stretch or shrink methods just before passing the images for model inference.

For the 'slices of interest' step \ref{step 1} training process we optimally needed to label all of the scan's slices that contain at least a part of an adrenal image, which is a lot of work for Radiologists to handle; on average we have 255 slices per scan, and on average 20\% of them are considered as slices of interest (see the Results chapter), so in total, a Radiologist should have labeled all 63750 ($1250 \cdot 255 \cdot 0.2$) images. That is of course impossible. Instead, a compromise was reached in which Radiologists labeled only a few slices from each scan containing an adrenal, then, for training, we considered all successive slices in between as slices of interest.

For the final scan classification task we labeled all slices with a seemingly \textit{abnormal} adrenal image, since the their prevalence is 10 times lower than a normal adrenal and it is much more important to have high quality labels for the abnormal classification's final step.

The dataset included 1250 scans, 122 of them were considered as being abnormal, each having at least one abnormal adrenal label. The left to right abnormal adrenal ratio was 1:5, resulting in a much smaller number of samples for a left abnormal adrenal. That is why we combined the left and right abnormal adrenal images for the classification task. Training an independent model for the left and right adrenals might result in a better performance for the right adrenal model, in which there is a sufficient number of cases, while  training a model for only the left abnormal adrenal label might have resulted in worse performance. Of course, in future, larger, data sets, this limitation can easily be overcome.

\subsection{Future Work}

In the following section we provide few directions for potential future work that can leverage our MMGA method.

\subsubsection{Integrate Reject Option - SelectiveNet}
The main idea is enabling the model to reject prediction on the hard cases that their confidence level is lower than others, this method is called by the author Yonatan Geifman "SelectiveNet"~\cite{geifman2019selectivenet}. This can give a big benefit to the system, by allowing the model to say "I don't know" for a predefined percentage of samples, by that the the accuracy improves and the tough cases will be directed to Radiologists attention to give more careful prediction, which the system can collect them for future learning on the hard cases.

The SelectiveNet is trained to optimize both classification (or regression) and rejection simultaneously (the rejection trained together with the entire network). A selective model is basically a pair $(f,g)$ where $f$ is a prediction function and $g:\chi \rightarrow \{0,1\}$ is a selection function, which serves as a binary qualifier for $f$ as follows:

\begin{equation}
(f,g)(x) =
\begin{cases}
f(x) & \text{if } g(x)=1 \\
\text{don't know} & \text{if } g(x)=0 \\
\end{cases}
\end{equation}

\subsubsection{Multi-modal learning - Combining CT and MRI scans}

A good direction is to examine how to combine different data types such as CT and MRI scans into a single model in a way that will harness the connections between CT and MRI so that eventually improve accuracy against a single scan type model. Existing studies in the field of automatically detecting pathologies from scans are based on a \textit{single} input type of images, such as CT scans, and do not consider other types of inputs such as MRI, due to the limited availability of paired MR-CT data. Thus, a good objective is to examine whether a system that can harness information from  paired CT-MR scans and perform a classification of a possible anomaly in the abdominal areas, might be better than a system using only one type of imaging.

\subsection{Conclusions}

\begin{enumerate}

\item Our MMGA method is provided as input  highly common CT scans that were originally directed and tuned for examining the spine, which were taken from patients with low back pain complaints. We use these CT scans for a quite different purpose, namely \textit{internal organs screening} -- automatically detecting totally different pathologies; in particular, we have demonstrated the methodology on the task of the detection of potentially suspicious lesions in the Adrenal glands.

\item Our method can support the detection of potential anomalies in other organs, such as the pancreas, the liver, or the kidneys; anomaly detection is not even restricted to the abdominal area and to CT scans and might also be useful for the analysis of abdominal, pelvic, and chest MRI scans, for example.

\item Our MMGA method can serve Radiologists in their daily routine as \textit{an automated screening tool} that can potentially  reduce the radiologists' workload and improve their diagnostic accuracy.  For example, it can be exploited for flagging abnormal Adrenals in spinal CT scans, as well as serve as another independent opinion that can strengthen radiologist's diagnosis.

\item Our method's evaluation demonstrates that it is possible to detect potential pathologies in multiple organs within a scan with multiple images, using one detection model architecture with a graph aggregation technique.

\item The MMGA compound pipeline architecture might be beneficial to other detection studies in the area of scanning videos or for the scanning of any entities containing multiple images.

\end{enumerate}

% \section{APPENDIX}
% \label{scn:Appendix}
% \input{Appendix}

% \cleardoublepage
% \phantomsection
% \addcontentsline{toc}{section}{\listfigurename}
% \addcontentsline{toc}{section}{\listtablename }
% \listoffigures
% \listoftables 

\nocite{*}

\bibliographystyle{abbrv}
\bibliography{bib}

\end{document}